\documentclass[letterpaper]{article} 
\usepackage{aaai2027}  
\usepackage[hyphens]{url}  
\usepackage{graphicx} 
\usepackage{natbib}  
\usepackage{caption} 
\usepackage{amsmath}
\usepackage{amssymb}
\usepackage{algorithm}
\usepackage{algorithmic}

\usepackage{booktabs}

\title{ResemBrick: Brick Reconstruction from Photographs with \\ Perceptual Fidelity and Buildability}
\author{
Xilun Chen\textsuperscript{\rm 4},
Hanwen Wan\textsuperscript{\rm 1,2},
Yusong Zhao\textsuperscript{\rm 4},
Zexin Lin\textsuperscript{\rm 1,2},
Ruixiang Liao\textsuperscript{\rm 1,2},
Xiaoqiang Ji\textsuperscript{\rm 1,2,3}\corresponding
} \affiliations{
\textsuperscript{\rm 1}School of Science and Engineering, The Chinese University of Hong Kong, Shenzhen, China\\
\textsuperscript{\rm 2}Shenzhen Institute of Artificial Intelligence and Robotics for Society, China\\
\textsuperscript{\rm 3}School of Artificial Intelligence, The Chinese University of Hong Kong, Shenzhen, China\\
\textsuperscript{\rm 4}School of Data Science, The Chinese University of Hong Kong, Shenzhen, China\\
\textsuperscript{*}jixiaoqiang@cuhk.edu.cn
}

\begin{document}

\nocopyright

\maketitle

\begin{abstract}

Producing a hand-buildable, colored brick model of a 3D object from a few casual photographs is a clean testbed for a broader challenge: generating 3D content that meets hard physical-assembly constraints under a discrete, budget-limited voxel grid. On a coarse lattice, visual resemblance and structural stability pull against each other, yet prior brick pipelines address only one side and treat voxelization as fixed preprocessing rather than a variable to optimize. We present \emph{ResemBrick}, which couples the two. Budgeted occupancy completion reframes discretization as allocation: given a target occupied-voxel count, a single resolution-conditioned network decides in one feed-forward pass which surface voxels to fill for best appearance, one weight set spanning 13 resolutions. Buildability by construction then combines support- and look-ahead-aware greedy placement with a deterministic, provably terminating repair that grounds every floating component. Under a matched budget, ResemBrick surpasses existing voxel selectors in perceptual fidelity while uniquely reaching zero floating and zero unstable bricks on unfiltered held-out objects; as a complete pipeline, it attains the best perceptual fidelity among prior brick-construction systems. Our results point to treating discretization and assembly as tightly coupled stages rather than independent ones.
\end{abstract}

\section{Introduction}
\label{sec:intro}

\providecommand{\methodname}{\emph{ResemBrick}}


\begin{figure}[t]
\centering
\includegraphics[width=\columnwidth]{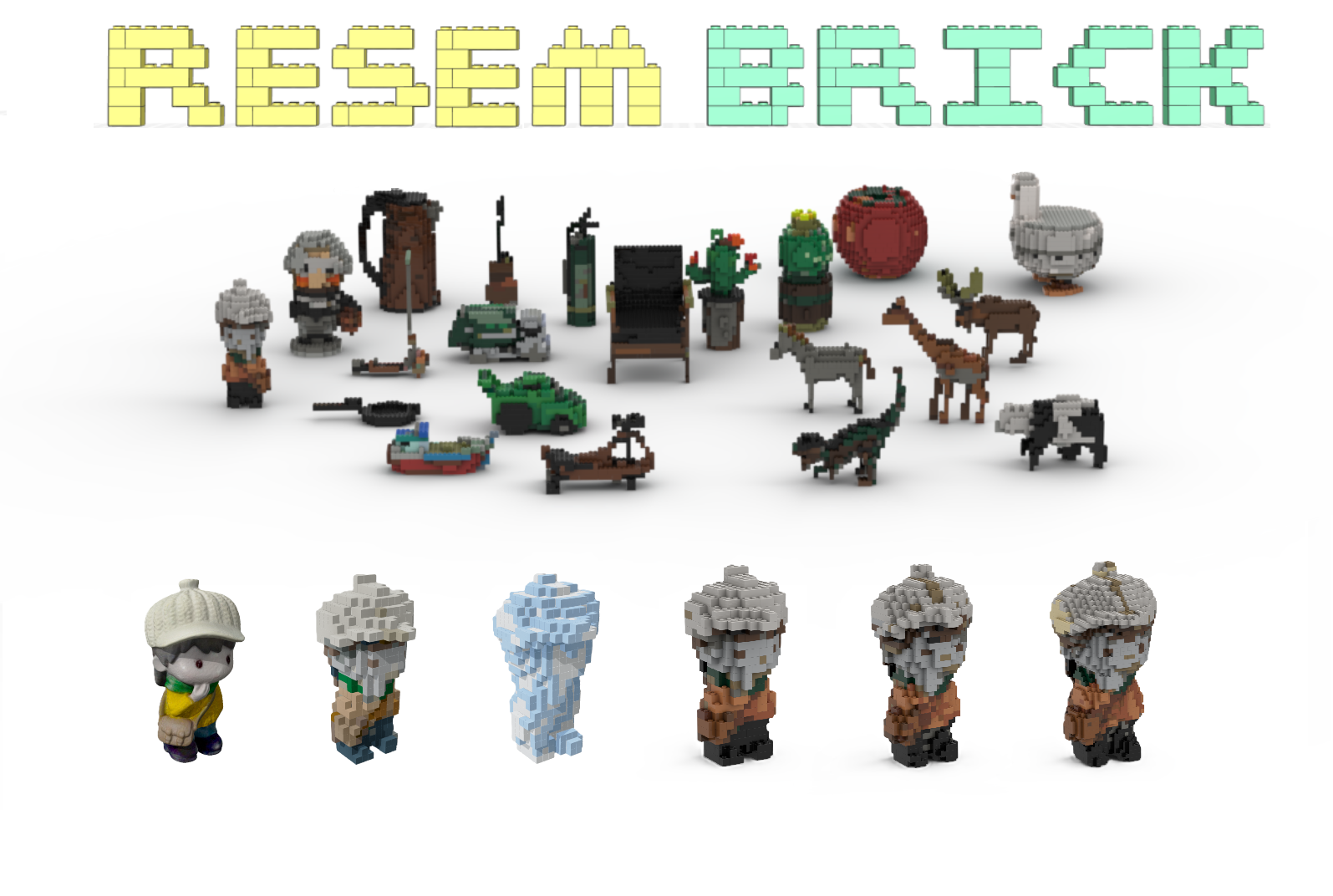}
\caption{Teaser.
Top: digital brick assemblies produced by \methodname{} from casual photographs. Bottom: one object across three resolutions: from left, input mesh, colored at $R{=}24$, budgeted-completion illustration, and the brick models at $R{=}24/32/64$.}
\label{fig:teaser}
\end{figure}

Generating a hand-buildable, colored brick model of a real-world object from pose-free multi-view photographs is a clean testbed for a broader problem: producing 3D content that satisfies hard physical-assembly constraints in a discrete-voxel setting. Recent pose-free multi-view reconstruction \cite{dust3r,freesplatter} makes recovering detailed geometry from a handful of casual photographs practical, while converting the mesh into a colored brick assembly still falls short.

Prior brick-modeling work tends to satisfy only one end. One line prioritizes perceptual or shape fidelity~\cite{silva2009,min2017,zhou2019} but ignores structural stability or is confined to a narrow object domain; another treats buildability as a physical-feasibility problem on a \emph{given} shape~\cite{testuz2013,luo2015}, or synthesizes structures from learned or generative priors~\cite{pun2025,stable2026,brickanything}, without aiming to faithfully reproduce appearance. As a result, faithful appearance and physical buildability are seldom guaranteed together within a single pipeline. Yet once bricks must be assembled for category-unconstrained objects, allocating discrete voxels on the coarse lattice is unavoidable. Under this coarse-lattice setting, the two goals pull against each other: spending the scarce occupied cells on shape-preserving structures may leave fragile, unbuildable geometry, while enforcing stability or uniformly thickening the voxel grid either flattens the thin structures that carry recognizability or over-inflates and distorts the overall shape. Budgeting voxel completion and brick assembly physical stability therefore cannot be handled in isolation and benefit from buildability-aware co-design: the former determines the upper bound attainable by the latter.

We therefore propose \methodname, which converts casual photographs of an everyday object into a physically buildable, colored brick assembly through two core designs, each resolving one side of this tension. \textbf{(i)~Budgeted occupancy completion} recasts low-resolution discretization as an allocation problem---under a user-specified number of occupied voxels, deciding which candidate surface voxels to fill to best preserve appearance, producing a low-resolution occupancy grid in a single feed-forward pass. \textbf{(ii)~Buildability by construction} combines support-, look-ahead-, and per-brick-color-aware greedy placement with a deterministic repair stage and a distilled stability surrogate that grounds any residual floating component.

Our contributions are threefold:
\begin{itemize}
 \item We introduce ResemBrick, a buildability-aware two-stage pipeline from pose-free photographs to buildable bricks that couples faithful appearance with grounded connectivity by construction.
 \item On the completion side, we recast voxelization as fidelity-driven budget allocation: a resolution- and budget-conditioned network, distilled from a buildability-repaired perceptual oracle, scores candidate voxels in one feed-forward pass. On the building side, we combine buildability-aware greedy placement with a deterministic, provably terminating repair stage that guarantees grounded connectivity and zero floating components.
 \item Experimentally, as a complete pipeline, ResemBrick attains the best perceptual fidelity and is the only one yielding fully stable assemblies against existing brick-construction systems; under a matched budget, its completion network further surpasses geometric voxel selectors in fidelity, confirming that budget allocation, not post-hoc assembly, sets the fidelity ceiling.
\end{itemize}


\section{Related Work}

\paragraph{Occupancy completion and budgeted voxelization.}
Voxelization is the design variable that prior brick pipelines mostly hold fixed. Classical voxelization fills cells according to geometric coverage or signed distance~\cite{nooruddin2003}, and existing brick systems inherit uniform or distance-based grids~\cite{pun2025,lennon2021}; this is reliable for coverage but indifferent to which voxels best preserve appearance under a coarse occupancy budget. Learned occupancy completion instead infers missing geometry, either as continuous fields~\cite{occnet} or as voxel completions from partial scans~\cite{voxelcompletion}, while recent discrete voxel methods optimize generated volumes for rendered appearance~\cite{voxify3d,dvd}. Our setting differs from both: the source mesh is already known, so the task is not to hallucinate unobserved structure but to decide which surface cells to occupy when the number of occupied cells is fixed to a target budget. We therefore cast low-resolution voxelization as budgeted allocation over a surface band and learn a selector that spends the limited occupancy where it most improves appearance resemblance, subject to a connectivity constraint that keeps the downstream assembly buildable.

\paragraph{Brick structure generation.}
Brick modeling has largely pursued either resemblance to a target shape or physical feasibility of a given assembly. Fidelity-oriented systems voxelize meshes or images into LEGO-like models while preserving color and contour: some target arbitrary objects~\cite{silva2009,min2017,lennon2021}, while others attain higher fidelity by specializing to a narrow object category such as architectural sculptures~\cite{zhou2019}, figurines~\cite{ge2024}, or micro buildings~\cite{ge2024_2}. Stability-aware methods instead optimize brick layouts, connectivity, or force balance on an already discretized shape~\cite{testuz2013,luo2015,peysakhov2003,bao2024,stablelego}; widely used authoring tools follow the same regime, focusing on constructability rather than preserving the object's appearance~\cite{bricklinkstudio}. Recent learned systems broaden the input and output space: BrickGPT~\cite{pun2025} generates brick structures autoregressively with rollback-based validity checks, BrickNet~\cite{bricknet} scales generation to large LDraw collections, and STABLE~\cite{stable2026} and BrickAnything~\cite{brickanything} drive buildability through learned or hand-designed rewards. Across these regimes, however, faithful appearance and physical stability are never guaranteed together for a low-budget reconstruction of an object.

\paragraph{Pose-free multi-view reconstruction.}
Our input side follows recent sparse-view 3D reconstruction. Neural radiance fields~\cite{mildenhall2020nerf} and 3D Gaussian Splatting~\cite{kerbl2023gaussiansplatting} established high-fidelity differentiable scene representations, typically relying on camera poses from structure-from-motion pipelines such as COLMAP~\cite{schoenberger2016sfm} and per-scene optimization. Generalizable Gaussian methods reduce this cost for sparse views~\cite{pixelsplat,mvsplat}, and pose-free reconstruction systems further remove calibration by predicting geometry, correspondences, Gaussians, and/or cameras from unposed images~\cite{dust3r,mast3r,noposplat,freesplatter}. We use FreeSplatter as a front end and extract a watertight textured mesh with TSDF fusion~\cite{curless1996volumetric} and Marching Cubes~\cite{lorensen1987marching}; any comparable reconstructor could substitute for it. The unresolved question, addressed here, is how to convert that recovered geometry into a low-budget brick assembly that both resembles the object and stands as built.

\section{Method}
\label{sec:method}

\subsection{Overview}
\label{sec:overview}

\begin{figure*}[t]
    \centering
    \includegraphics[width=\textwidth]{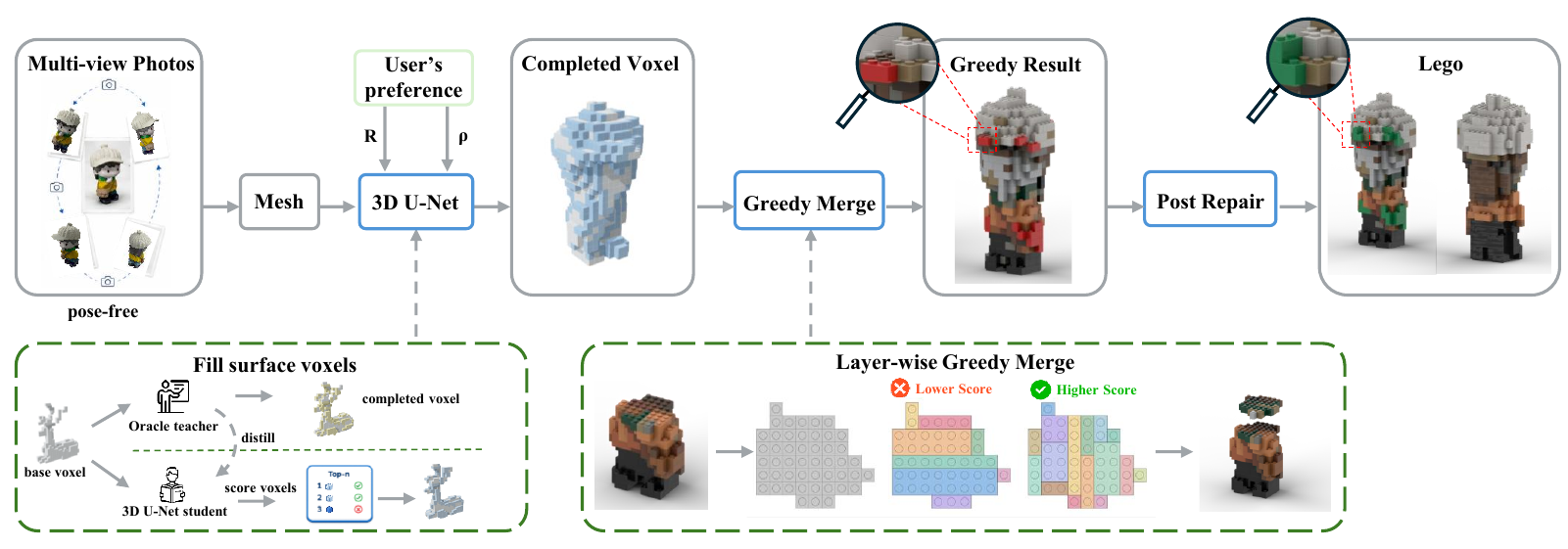}
    \caption{End-to-end image-to-buildable-brick pipeline.
    Two stages meet at a low-resolution colored occupancy grid:
    the voxelization stage (photo$\rightarrow$voxel) and the building
    stage (voxel$\rightarrow$brick).}
    \label{fig:pipeline}
\end{figure*}

Given a handful of casual multi-view photographs of an object, we produce a physically buildable, colored brick model (Fig.~\ref{fig:pipeline}). A pose-free reconstructor (FreeSplatter~\cite{freesplatter}) lifts the input views into a set of Gaussians and camera poses, from which TSDF fusion and Marching Cubes recover a watertight textured mesh $\mathcal{M}$.

The voxelization stage is a single resolution-transferable network which selects which surface voxels best preserve appearance while keeping a stable assembly feasible under a fixed budget (Sec.~\ref{sec:stageA}). The building stage partitions each layer's 4-connected region into non-overlapping library bricks while enforcing inter-layer grounded connectivity, a combinatorially large search related to rectangle partition~\cite{lingas1982power}. We handle it with a buildability-aware greedy procedure whose repair step guarantees grounded connectivity (Sec.~\ref{sec:stageB}).


\subsection{Voxelization Stage: Buildability-Aware Occupancy Completion}
\label{sec:stageA}

The reconstructed mesh $\mathcal{M}$ is normalized to $[0,1]^3$. The naive uniform voxelization spends budget on structurally stable regions while neglecting the structurally fragile, perceptually important thin structures. We instead treat voxelization as \emph{budgeted occupancy completion} at a fixed low resolution $K$: the unambiguous deep-interior and far-exterior voxels are fixed, and learning is restricted to the \emph{surface band} $\mathcal{S}$, the shell of voxels straddling the surface where the fill decision is nontrivial. Given a user-controlled \emph{target} density $\rho_{\mathrm{tgt}}\in[0,1]$, the fill count is set geometrically as $n_{\mathrm{occ}}(\rho_{\mathrm{tgt}}) = n_{\mathrm{base}} + \mathrm{round}(|\mathcal{S}|\rho_{\mathrm{tgt}})$, with $n_{\mathrm{base}}$ the uniform-baseline count. We complete the band with a fully-convolutional 3D U-Net~\cite{ronneberger2015unet,cicek20163dunet} with skip connections, conditioned on the target resolution and the \emph{realized} density $[R,\rho]$ via FiLM~\cite{perez2018film} so that one set of weights serves $R\in\{24,32,48\}$. As occupancy is discrete, a gap arises between the soft scores seen during training and the thresholding applied at inference; we therefore train in two phases: first distilling an optimized completion oracle, then refining it against rendered quality. At inference the user supplies a target density $\rho_{\mathrm{tgt}}$, and we take the top-$n_{\mathrm{occ}}(\rho_{\mathrm{tgt}})$ scored band voxels as the completed grid. The completed grid finally inherits its appearance from the textured mesh by barycentric color sampling and a diversity-aware CIELAB palette; details are in the supplementary material.

\subsubsection{Training phase 1: Oracle Distillation}
\label{sec:stageA1}

The \emph{oracle} is an offline-constructed supervision target. It selects, from the surface band $\mathcal{S}$, $n_{\mathrm{occ}}(\rho_{\mathrm{tgt}})$ voxels whose filling best reconstructs the object, greedily adding the band voxel that most reduces a geometry-aware perceptual loss $\mathcal{Q}$, measured between six-view renderings of the candidate grid and the target and aggregating depth, silhouette, and normal agreement,

\begin{equation}
\label{eq:oracle_q}
\begin{aligned}
\mathcal{Q} = {}& \lambda_d\,\|\hat{d}-d\|_1
            + \lambda_a\,\|\hat{\alpha}-\alpha\|_1 \\
          & + \lambda_n\bigl(1-\cos(\hat{\mathbf{n}},\mathbf{n})\bigr)
            + \lambda_p\,\mathrm{LPIPS}(\hat{d}, d),
\end{aligned}
\end{equation}
where $d,\alpha,\mathbf{n}$ are the rendered depth, silhouette (coverage) mask, and surface normals, and hats denote the candidate grid. We evaluate LPIPS on the \emph{colorless depth} rather than on textured renderings: on brick/voxel grids the hard color seams between adjacent cells dominate a natural-image perceptual metric and yield spurious scores, whereas depth isolates how well the occupancy fits mesh boundary. Because each greedy step re-scores only voxels whose neighborhood changed, we use lazy (CELF-style) greedy evaluation with batched six-view rendering, making the per-object oracle construction tractable. The selection is run at three fill levels $\rho_{\mathrm{tgt}}\in\{0.4,0.62,0.75\}$ per object.

The greedily selected grid maximizes fidelity but need not be physically buildable, so each candidate is screened and repaired. Objects whose center of mass does not project within the convex hull of their ground-contact footprint are discarded as statically unstable. The rest are repaired to a single \emph{6-connected} component, by keeping the largest component as the trunk and bridging each floating component to its nearest trunk along an $L_1$ stair path within a gap tolerance, or removing it otherwise. 

Because repair alters the filled count, each grid is re-labeled with its post-repair density $\rho$, and the U-Net is supervised on these final band labels with a combined objective,
\begin{equation}
\label{eq:s1loss}
\mathcal{L}_{\mathrm{vox1}}
  = \mathrm{BCE}\!\left(\hat{y}, y\right)
  + \mathrm{Dice}\!\left(\hat{y}, y\right),
\end{equation}
whose Dice term counters the band's positive-rate imbalance and desaturates the score distribution for the next stage to refine. At inference $\rho$ remains a user-control input.

\subsubsection{Training phase 2: Annealed STE Refinement}
\label{sec:stageA2}

Phase 2 refines the network directly against rendered quality under a straight-through estimator (STE)~\cite{bengio2013estimating}: the forward pass binarizes the soft scores to hard occupancy for rendering, while the backward pass treats the threshold as the identity so gradients reach the soft scores. The refinement objective combines perceptual fidelity (LPIPS and SSIM, evaluated on colorless depth renderings) with the budget constraint,

\begin{equation}
\label{eq:s2loss}
\begin{aligned}
\mathcal{L}_{\mathrm{vox2}}
  = {}& w_{\ell}\,\mathrm{LPIPS} + w_{s}\,(1-\mathrm{SSIM}) \\
     & + w_{b}\,\bigl(n_{\mathrm{filled}}-n_{\mathrm{occ}}\bigr)^2 ,
\end{aligned}
\end{equation}

evaluated on renderings of the hard grid against the target. The budget term prevents the degenerate all-fill/all-empty solutions. The STE temperature $\tau$ is annealed geometrically from $2.0$ to $0.1$, so training conditions increasingly resemble deployment. Although Phase 2 optimizes only perceptual fidelity, it preserves the buildability inherited from Phase 1: floating and ungrounded voxel rates stay below $0.05\%$ and the unsupported rate is essentially unchanged (Sec.~\ref{sec:exp_stageA}).


\subsection{Building Stage: Buildability-Aware Greedy Brick Assembling}
\label{sec:stageB}

The building stage partitions the colored occupancy grid into standard bricks, splitting each layer into 4-connected components processed independently. Buildability is handled in two complementary ways: as soft preferences during greedy scoring, and as a hard guarantee from a subsequent repair stage. Algorithm~\ref{alg:merge} summarizes the procedure.

\begin{algorithm}[t]
\caption{Buildability-Aware Greedy Brick Assembling}
\label{alg:merge}
\begin{algorithmic}[1]
\REQUIRE Colored occupancy $\mathcal{V}$; max bridge gap $d_g$
\ENSURE Grounded, connected brick set $\mathcal{B}$
\STATE $\mathcal{B} \gets \emptyset$
\FORALL{4-connected components $R$ over all layers}
    \WHILE{$R \neq \emptyset$ and large candidates remain}
        \STATE $\mathcal{C} \gets$ library rectangles fitting $R$
        \STATE score each $b \in \mathcal{C}$ by Eq.~\eqref{eq:score}
        \STATE $\mathcal{C}^{*} \gets$ non-overlapping top-scoring subset
        \STATE $\mathcal{B} \gets \mathcal{B} \cup \mathcal{C}^{*}$;\quad $R \gets R \setminus \bigcup_{b \in \mathcal{C}^{*}} F(b)$
    \ENDWHILE
    \STATE fill $R$ with $1{\times}1$ bricks (supported cells first)
    \STATE merge adjacent bricks into library rectangles; absorb leftovers
\ENDFOR
\STATE \COMMENT{Global floating-component repair}
\REPEAT
    \STATE build union--find;\quad $\mathcal{S}_F \gets$ floating components
    \STATE pick $\mathcal{S}_K \in \mathcal{S}_F$
    \STATE $\sigma^{*} \gets$ first success in ordered list: zero-deformation
    \STATE \quad (re-merge, re-cut, re-bridge), then voxel-adding (cap, shelf)
    \IF{$\sigma^{*}$ verified (floating count decreases)}
        \STATE apply $\sigma^{*}$
    \ELSE
        \STATE remove $\mathcal{S}_K$
    \ENDIF
\UNTIL{$\mathcal{S}_F = \emptyset$}
\RETURN $\mathcal{B}$
\end{algorithmic}
\end{algorithm}

\subsubsection{Candidate Scoring.}
\label{sec:stageB_score}
Starting from a component of unit cells, the algorithm repeatedly scores every candidate brick $b$ (a library rectangle that fits the remaining region $R$) and greedily commits a non-overlapping set of top-scoring bricks. The score balances coverage, buildability, and color fidelity against a look-ahead penalty:
\begin{equation}
\label{eq:score}
\begin{aligned}
S(b\mid R) = {}&
  w_a\,|F(b)|
  + \Phi_{\mathrm{sup}}(b) \\
  & + w_c\,\Phi_{\mathrm{col}}(b)
  - \Phi_{\mathrm{fut}}(b\mid R).
\end{aligned}
\end{equation}
The first term rewards large footprints $|F(b)|$, favoring fewer, larger bricks.

\paragraph{Buildability-aware support ($\Phi_{\mathrm{sup}}$).}
The support term is the core of our buildability-aware scoring. Let $r(b)\in[0,1]$ be the fraction of $F(b)$ solidly supported by the layer below and $n_{\mathrm{un}}(b)$ the number of unsupported cells the brick covers. Ground-layer bricks are fully supported; otherwise
\begin{equation}
\label{eq:support}
\begin{aligned}
\Phi_{\mathrm{sup}}(b) = {}&
  w_s\,|F(b)|\,r(b) \\
  & + \underbrace{w_s\,n_{\mathrm{un}}(b)\,
    \mathbf{1}\!\left[0 < r(b) < 1\right]}_{\text{rescue bonus}} .
\end{aligned}
\end{equation}
The first part rewards bricks on solid support. The \emph{rescue bonus} activates when a brick simultaneously covers supported and unsupported cells: such a brick spans an overhang, anchoring otherwise floating cells to the structure below.

\paragraph{Color fidelity ($\Phi_{\mathrm{col}}$).}
Because a physical brick carries a single color, spanning cells of different colors misrepresents all but one. The color term $\Phi_{\mathrm{col}}$ weights each such mismatch by an importance--rarity salience, which rises for locally isolated colors and for globally rare ones, so brick seams favor color edges over uniform regions (full expressions in the supplementary material).

\paragraph{Look-ahead penalty ($\Phi_{\mathrm{fut}}$).}
The look-ahead penalty discourages placements that strand cells for later steps, counting residual cells that lack support from below and have fewer than $k$ lateral neighbors, and weighting interior more than boundary cells,
\begin{equation}
\label{eq:future}
\begin{aligned}
\Phi_{\mathrm{fut}}(b\mid R)
  = {}& \lambda_{\mathrm{int}}\,U_{\mathrm{int}}
  + \lambda_{\mathrm{edge}}\,U_{\mathrm{edge}}, \\
  & \lambda_{\mathrm{edge}} < \lambda_{\mathrm{int}},
\end{aligned}
\end{equation}
as improper interior placements are more likely to leave fragile, unsupported cells.

After greedy placement, residual cells are filled with $1{\times}1$ bricks (supported cells first to preserve merge opportunities), hierarchically merged into valid library rectangles in ascending target area, and any leftover $1{\times}1$ brick is absorbed into a library-valid neighbor.

\subsubsection{Post-Merge Stability.}
\label{sec:stageB_stability}
\begin{figure}[t]
\centering
\includegraphics[width=\columnwidth]{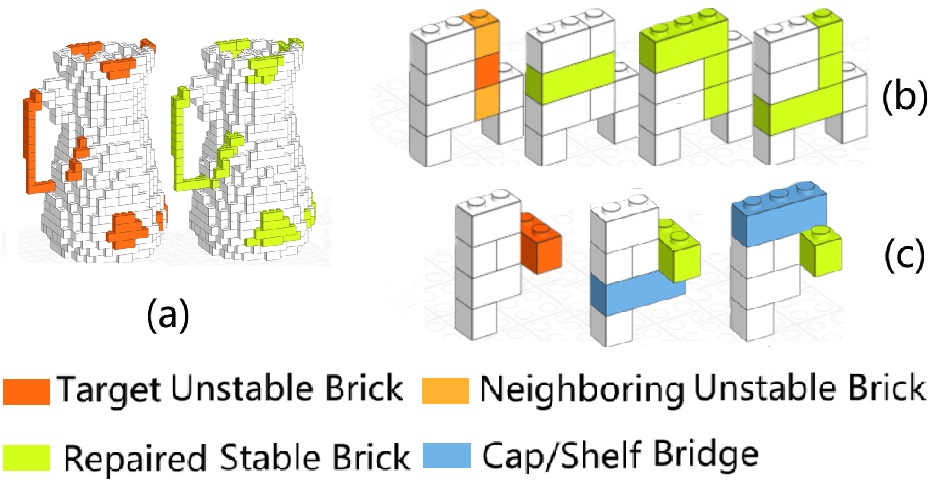}
\caption{Two-tier floating repair. (a)~Demonstration of one object's floating and fixed components. Zero-deformation
recombination (b) is preferred; voxel-adding bridges (c) are a fallback.}
\label{fig:repair}
\end{figure}

Merging may leave brick groups disconnected from the ground. A union--find structure flags any \emph{floating} component, which the repair operators in Algorithm~\ref{alg:merge} resolve in a two-tier order: zero-deformation recombination first, including re-merging or re-cutting neighboring bricks into a spanning brick anchored by its supported segment; then voxel-adding cap/shelf bridges as a fallback (Fig.~\ref{fig:repair}). Each attempt is atomic, so the assembly reaches full ground connectivity in finitely many steps. When no repair operator succeeds, the offending floating component is removed as a last resort (Algorithm~\ref{alg:merge}, line~21). In practice, this branch fires rarely and removes a negligible voxel fraction (Sec.~\ref{sec:exp_stageB_removal}). Global static stability is then certified on the finished assembly by a lightweight message-passing network over the assembly graph (bricks as nodes, vertical contacts as edges), which is an encode--process--decode architecture with $M$ message-passing layers and a dual head (per-brick stability verdict and margin) trained on StableLego's~\cite{stablelego} per-brick stability labels from force-balance analysis. Its dimensionless features transfer across voxel resolutions, and it emits a per-brick verdict at millisecond cost, reproducing the exact force-balance solver at a per-brick AUC of $0.984$; architecture and full distillation results are in the supplementary material.

\section{Experiments}
\label{sec:experiments}
\subsection{Experimental Setup}
\label{sec:exp_setup}
\paragraph{Data and protocol.}
Our voxelization network is trained and evaluated on OmniObject3D~\cite{omniobject3d}, a large collection of high-quality scanned single objects. We split by instance into $3070$ training and $198$ validation objects (one held-out instance per category); the same $198$-object validation set is used throughout the voxelization, building, and full-pipeline evaluations. The building stage is additionally evaluated on StableText2Brick~\cite{stabletext2lego}, a clean buildability benchmark. All full-pipeline and buildability comparisons on OmniObject3D are conducted at $R{=}24$, the coarsest resolution: this is the largest lattice at which the exact force-balance solver and the Gurobi-based baselines remain tractable within our compute budget. BrickGPT's per-brick Gurobi stability check, which we leave unmodified, times out repeatedly at finer resolutions. All voxelization comparisons operate under the matched budget $n_{\mathrm{occ}}(\rho)$: baselines are derived by construction from the same frozen SDF, camera poses, and mesh. Significance is assessed with per-object paired Wilcoxon signed-rank tests against Ours ($^{*}p{<}.05$, $^{**}p{<}.01$, $^{***}p{<}.001$).

\paragraph{Metrics.}
Perceptual fidelity is our primary criterion, reported as MS-SSIM~\cite{wang2003msssim} and LPIPS~\cite{zhang2018lpips} (AlexNet) on rendered views. Geometry serves as a guardrail rather than the objective: we report Chamfer distance~\cite{fan2017pointset} and F-score~\cite{tatarchenko2019singleview} ($\tau{=}0.02$, unit-cube normalization). All reported stability numbers (\emph{Stab\%}) are certified by the exact Gurobi force-balance solver. The full metric set (PSNR, SSIM~\cite{wang2004ssim}, DISTS~\cite{ding2020dists}, VGG-LPIPS, normal consistency) is given in the supplement. Buildability metrics are specific to each stage and defined where they are first used.

\paragraph{Overview.}
We structure the experiments to answer the following questions:
\begin{itemize}
\item[\textbf{Q1}] Does ResemBrick's overall performance beat existing mesh-to-brick systems on fidelity and stability?(Sec.~\ref{sec:exp_e2e})
\item[\textbf{Q2}] Is ResemBrick's output buildable by hand in the real world? (Sec.~\ref{sec:exp_realworld})
\item[\textbf{Q3}] Is the completion network's architecture necessary and capacity-efficient? (Sec.~\ref{sec:exp_stageA_ablation})
\item[\textbf{Q4}] Under a matched budget, does our completion beat geometric selectors? (Sec.~\ref{sec:exp_stageA_sel})
\item[\textbf{Q5}] Does the conditioned network generalize across resolutions? (Sec.~\ref{sec:exp_stageA_res})
\item[\textbf{Q6}] Does our building stage produce more stable, fewer-brick assemblies? (Sec.~\ref{sec:exp_stageB})
\item[\textbf{Q7}] Does the repair achieve zero floating at a negligible cost to shape? (Sec.~\ref{sec:exp_stageB_removal})
\end{itemize}

\subsection{Overall Performance of ResemBrick}
\label{sec:exp_e2e}
\begin{figure*}[t]
\centering
\includegraphics[width=\textwidth]{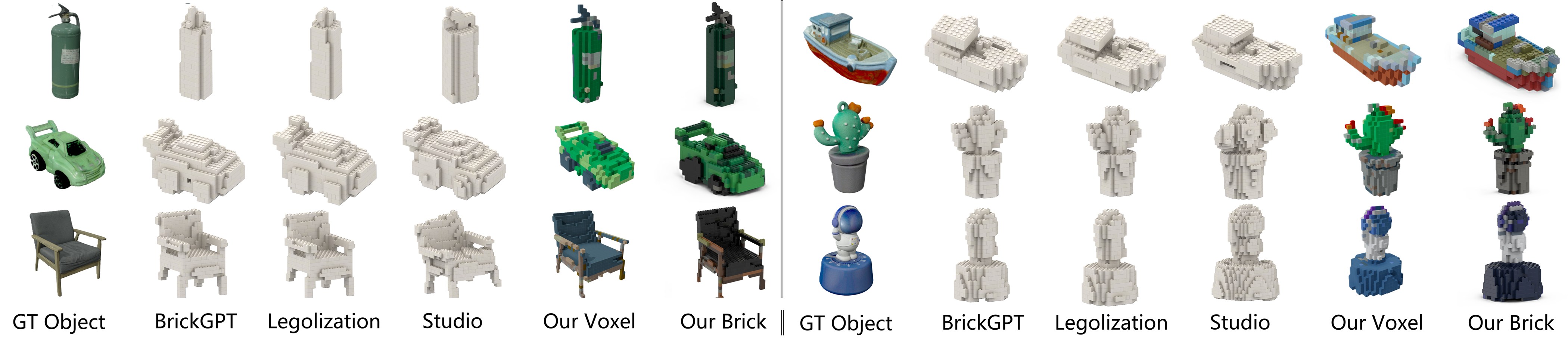}
\caption{End-to-end gallery. Six objects (two per row); for each,
left to right: input mesh, brick assemblies from BrickGPT, Legolization, and BrickLink Studio,
our completed colored voxels, and our pipeline.}
\label{fig:e2e_gallery}
\end{figure*}

We evaluate the full pipeline from photographs to buildable brick models by chaining reconstruction, voxelization, and building. The comparison uses the held-out OmniObject3D set ($198$ objects) at $R{=}24$, with this resolution choice described in the setup. We compare with three accessible brick-modeling systems: BrickGPT~\cite{pun2025}, Legolization~\cite{luo2015}, and the commercial converter BrickLink Studio~\cite{bricklinkstudio}. Ours attains the best perceptual fidelity while being the only pipeline whose assemblies are fully stable (Table~\ref{tab:e2e}, Fig.~\ref{fig:e2e_gallery}).

\begin{table}[t]
\centering
\smallskip
\small
\begin{tabular}{l c c c}
\toprule
\textbf{Method} & MS-SSIM$\uparrow$ & LPIPS$_{\mathrm{a}}\downarrow$ & Stab\%$\uparrow$ \\
\midrule
BrickGPT & 0.940 & 0.081 & 82.7 \\
Legolization & 0.940 & 0.082 & 70.6 \\
BrickLink Studio & 0.733 & 0.204 & 75.0 \\
\midrule
\textbf{Ours} & \textbf{0.955} & \textbf{0.075} & \textbf{100} \\
\bottomrule
\end{tabular}
\caption{Full pipeline on the held-out OmniObject3D set
($198$ objects, $R{=}24$). Headline perceptual metrics (MS-SSIM $\uparrow$,
LPIPS$_{\mathrm{a}}$ $\downarrow$) and buildability (\emph{Stab\%}:
fraction of assemblies with zero floating and zero
statically unbalanced bricks under an exact solver, $\uparrow$). Best in bold.}
\label{tab:e2e}
\end{table}

\subsubsection{Real-world validation.}
\label{sec:exp_realworld}

\begin{figure}[t]
\centering
\includegraphics[width=\columnwidth]{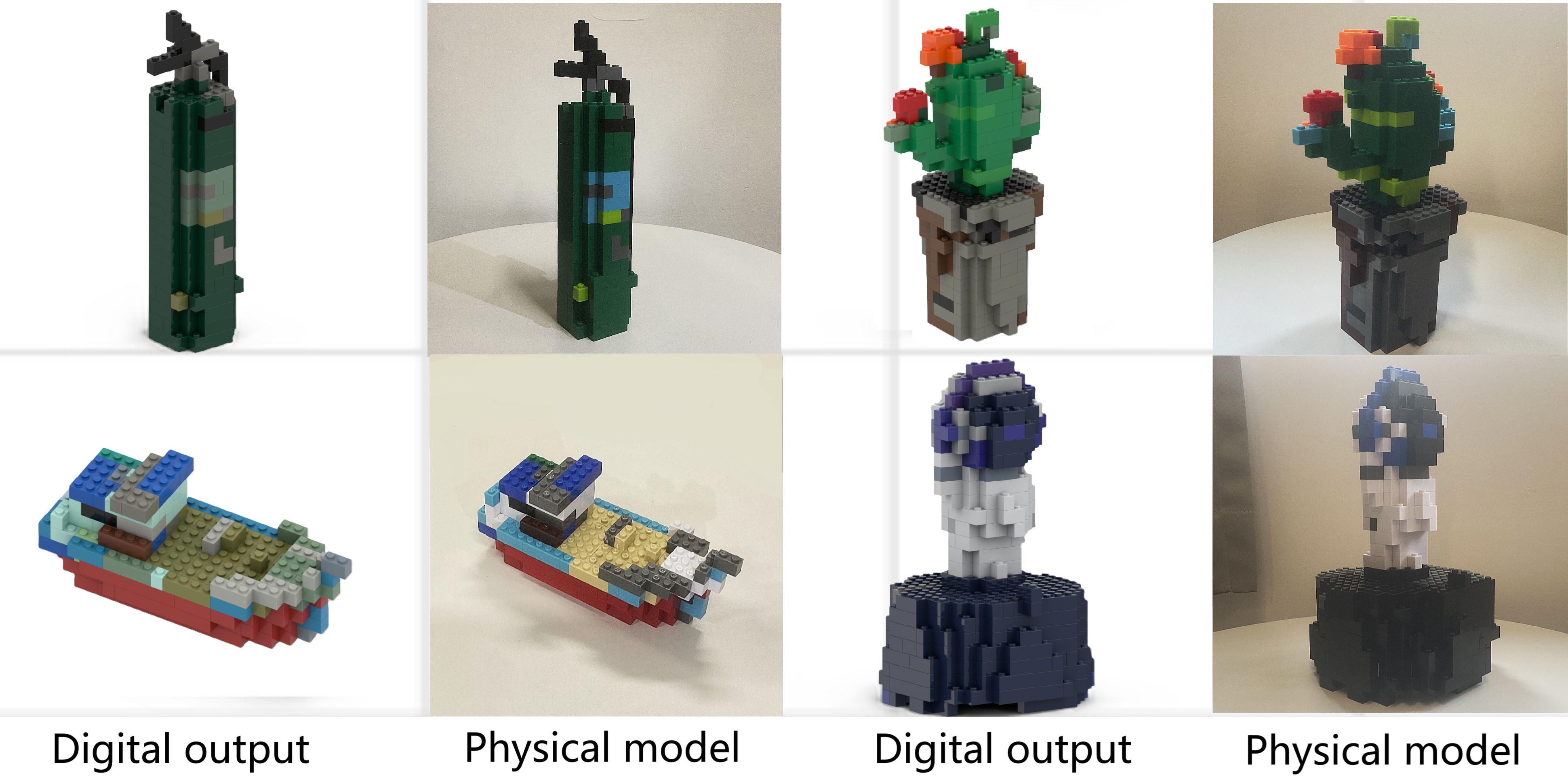}
\caption{Real-world validation. Four objects built by hand from the pipeline's exported layer-by-layer instructions; digital output (left) vs.\ physical model (right). Minor color differences reflect same-size substitutions where a color is unavailable for the required brick.}
\label{fig:realworld}
\end{figure}

We physically build four objects spanning diverse categories by hand, following the exported layer-by-layer instructions directly from the pipeline without manual edits (Fig.~\ref{fig:realworld}). Every model stands unsupported on a flat surface and reproduces the digital shape and colors, confirming that the connectivity constraints and post-repair translate into real physical stability. Minor color differences arise where a color is unavailable in the required brick size, in which case we substitute the closest same-size color.

\subsection{Voxelization Stage}
\label{sec:exp_stageA}

We assess the voxelization stage in three steps: we first validate the completion network's design, then compare it against alternative selectors under a matched occupancy budget $n_{\mathrm{occ}}(\rho)$, and finally test its generalization across resolutions. The comparison pits our learned completion (Ours) against five purely geometric selectors and a TSDF-fusion reference that additionally consumes ground-truth depth. Therefore, TSDF served as an oracle-input upper bound rather than a competing selector. Beyond fidelity, we evaluate buildability by the number of connected components (\#Comp) and the fraction of floating and unsupported voxels.

\subsubsection{Architecture and capacity ablation.}
\label{sec:exp_stageA_ablation}
The completion network is a fully-convolutional 3D U-Net (channel widths $8{\to}16{\to}32$, two down/up-sampling stages, skip connections, GroupNorm layers~\cite{wu2018group}, and a zero-initialized output layer), with the density--resolution condition $[R,\rho]$ injected through FiLM at every convolutional block. A single set of weights ($0.10$M parameters) serves all resolutions; full hyperparameters are in the supplementary material.

These design choices are deliberate (Table~\ref{tab:ablation}). The architecture and capacity choices in Table~\ref{tab:ablation} were made on the basis of training loss and early stopping together with a parameter-count prior, not on the $198$-object evaluation set, which is reserved exclusively for the final evaluation. The conditioning-injection scheme is not load-bearing, group normalization is essential, and skip connections help modestly. Crucially, capacity is not the bottleneck: our default $0.10$M model matches models up to $57{\times}$ larger.

\begin{table}[t]
\centering\small
\setlength{\tabcolsep}{4pt}
\begin{tabular}{lc @{\hspace{1.4em}} lcc}
\toprule
\multicolumn{2}{c}{Architecture} & \multicolumn{3}{c}{Capacity} \\
\cmidrule(r){1-2}\cmidrule(l){3-5}
Variant & IoU & Width & Par. & IoU \\
\midrule
FiLM (ref.)      & $0.657$        & $8$ (def.) & $0.10$M & $0.653$ \\
concat           & $0.655$        & $16$       & $0.38$M & $0.654$ \\
cross-attn       & $0.651$        & $24$       & $0.83$M & $0.656$ \\
FiLM @bottleneck & $0.657$        & $32$       & $1.45$M & $0.657$ \\
no GroupNorm     & \emph{fails}   & $48$       & $3.22$M & $0.657$ \\
no skip          & $0.647$        & $64$       & $5.68$M & $0.656$ \\
\bottomrule
\end{tabular}
\caption{Architecture ablation (left) and
capacity ablation (right), val IoU vs.\ the completion oracle
(avg). \emph{no GroupNorm} does not train (IoU frozen at initialization,
$0.620$).}
\label{tab:ablation}
\end{table}

\subsubsection{Completion vs.\ geometric selectors.}
\label{sec:exp_stageA_sel}

Ours is both the most faithful and the most buildable learned grid (Tables~\ref{tab:stagea_fidelity},~\ref{tab:stagea_buildability}): it wins perceptual fidelity against all geometric selectors while matching the strongest on the geometry guardrail, and attains the fewest connected components with near-zero floating voxels. Post-hoc repair does not cost appearance: it bridges only ${\sim}0.01\%$ of voxels, leaving fidelity unchanged while restoring full connectivity (\#Comp${=}1$). Finally, the TSDF control shows that buildability must be targeted during training rather than assumed: even fed ground-truth depth, naive fusion yields markedly less buildable structure than our budgeted completion.

\begin{table}[t]
\centering\small
\setlength{\tabcolsep}{5pt}
\begin{tabular}{lcccc}
\toprule
Method & LPIPS$_\mathrm{a}\downarrow$ & MS-SSIM$\uparrow$
& Chamfer$\downarrow$ & F-score$\uparrow$ \\
\midrule
SDF shell        & 0.110*** & 0.930*** & 0.045*** & 0.523*** \\
Top-$K$          & 0.103*** & 0.923*** & \textbf{0.037}*** & \textbf{0.654}*** \\
Dilate           & 0.104*** & 0.922*** & 0.038*** & 0.653*** \\
6-sep            & 0.119*** & 0.907*** & 0.040*** & 0.645 \\
26-sep           & 0.120*** & 0.881*** & 0.045*** & 0.610*** \\
TSDF$^\ddagger$  & 0.090*** & 0.946*** & 0.030*** & 0.735*** \\
\midrule
\textbf{Ours}    & \textbf{0.098} & \textbf{0.932} & 0.038 & 0.641 \\
+repair$^\dagger$ & 0.099 & 0.932 & 0.038 & 0.641 \\
\bottomrule
\end{tabular}
\caption{Voxelization fidelity under matched budgets $n_{occ}(\rho)$
(avg.\ over $\rho\in\{0.4,0.62,0.75\}$, $R\in\{24,32,48\}$). Perceptual:
LPIPS$_\mathrm{a}$, MS-SSIM; geometry guardrail: Chamfer, F-score
($\tau{=}0.02$). TSDF$^\ddagger$ is an oracle-input reference
(ground-truth depth), excluded from the \textbf{bold} best.}
\label{tab:stagea_fidelity}
\end{table}

\begin{table}[t]
\centering\small
\setlength{\tabcolsep}{5pt}
\begin{tabular}{lccc}
\toprule
Method & \#Comp$\downarrow$ & Float.\%$\downarrow$ & Unsup.\%$\downarrow$ \\
\midrule
SDF shell        & 2.12*** & 0.30*** & 4.41*** \\
Top-$K$          & 1.78*** & 0.02*** & 3.32 \\
Dilate           & 1.76*** & 0.09*** & 3.31 \\
6-sep            & 3.25*** & 0.43*** & 3.56 \\
26-sep           & 2.34*** & 2.38*** & 3.48* \\
TSDF$^\ddagger$  & 1.92*** & 0.33*** & 3.72** \\
\midrule
\textbf{Ours}    & 1.61 & \textbf{0.03} & \textbf{3.14}*** \\
+repair$^\dagger$ & \textbf{1.00} & \textbf{0.00} & 3.52 \\
\bottomrule
\end{tabular}
\caption{Voxelization buildability under matched budgets $n_{occ}(\rho)$
(avg.\ over $\rho\in\{0.4,0.62,0.75\}$, $R\in\{24,32,48\}$). \#Comp:
connected components; Float./Unsup.: floating/unsupported voxels;
$^\dagger$after repair the budget is no longer exactly $n_{occ}$.}
\label{tab:stagea_buildability}
\end{table}

\subsubsection{Resolution generalization.}
\label{sec:exp_stageA_res}
\begin{figure}[t]
\centering
\includegraphics[width=\columnwidth]{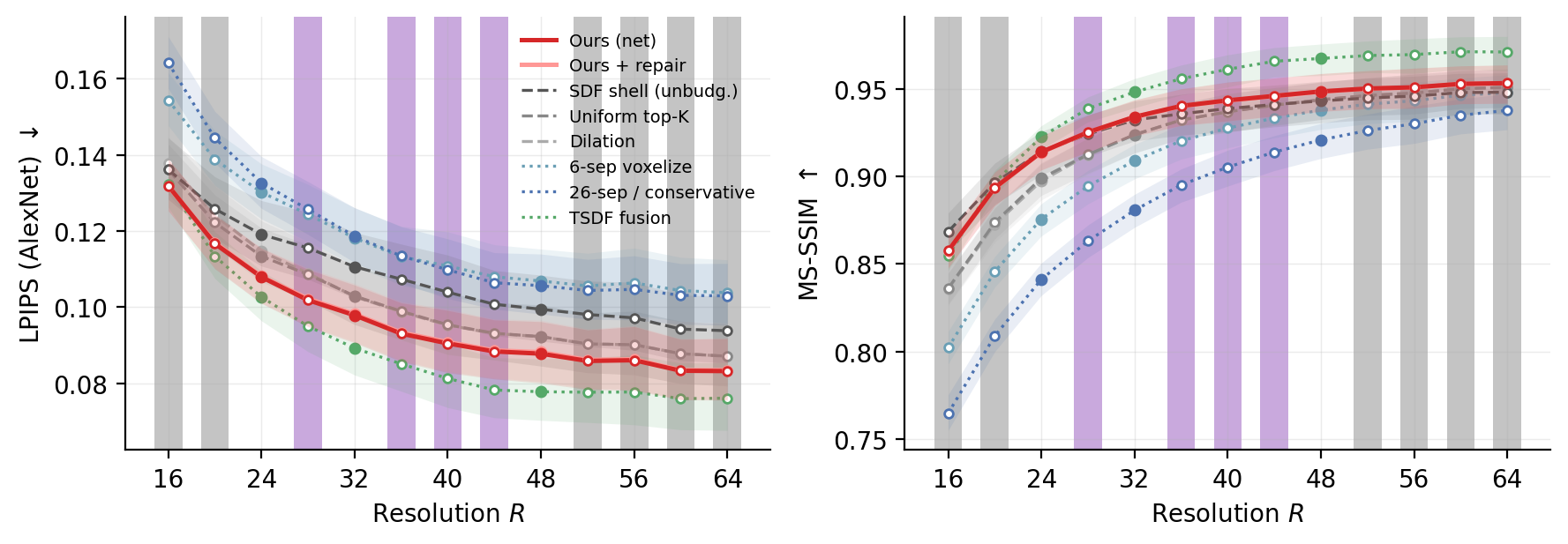}
\caption{Resolution generalization at $\rho{=}0.62$.
LPIPS$_{\mathrm{a}}$ ($\downarrow$) and MS-SSIM ($\uparrow$) across
$13$ resolutions. Filled markers are the trained resolutions
$\{24,32,48\}$; open markers are unseen (interpolated $28/36/40/44$;
extrapolated $16/20$ and $52$--$64$). Shaded bands are $95\%$ bootstrap
CIs.}
\label{fig:stagea_multireso}
\end{figure}

A single conditioned network handles the full range without retraining (Fig.~\ref{fig:stagea_multireso}), leading on LPIPS$_{\mathrm{a}}$ across all $13$ resolutions and on MS-SSIM for $R\ge28$; at the coarsest grids ($R\le24$) it trails only the sparse SDF shell, as structural-similarity metrics favour sparse shells on very coarse grids. The advantage holds at both interpolated and extrapolated resolutions, indicating that the FiLM conditioning learns a continuous map over resolution rather than memorizing discrete settings.

\subsection{Building Stage}
\label{sec:exp_stageB}
We compare against brick-construction methods with a publicly available implementation---BrickGPT, Legolization, and STABLE~\cite{stable2026}---all on the same grid over StableText2Brick (\textsc{s2b}) and the held-out OmniObject3D subset (\textsc{omni}). Table~\ref{tab:stageB} (bottom) ablates each component.

Ours uses the fewest bricks among (near-)fully stable methods while remaining (near-)fully buildable on both sets, and is the only method with zero floating and zero statically unbalanced bricks on the harder held-out \textsc{omni}. The ablation confirms every term contributes: removing repair collapses stabilityon s2b(52.4\%), and reintroduces floating/unstable bricks on omni, while support and look-ahead further trim brick count and fragmentation.

\paragraph{Repair cost.}
\label{sec:exp_stageB_removal}
Zero-floating guarantee comes at a negligible cost to shape. On the held-out \textsc{omni} set, the removal fallback fires on only $23$ of $198$ objects and discards $57$ of $61{,}078$ bricks ($0.093\%$; $0.025\%$ of voxels), while the bridging operators add $0.104\%$ of voxels; the two roughly cancel, leaving net coverage slightly \emph{positive} ($+0.079\%$). 

\begin{table}[t]
\centering\small
\setlength{\tabcolsep}{1pt}
\smallskip
\begin{tabular*}{\columnwidth}{@{\extracolsep{\fill}}l c c c c c c c c@{}}
\toprule
& \multicolumn{4}{c}{\textsc{s2b} ($42$ obj.)} &
\multicolumn{4}{c}{\textsc{omni} ($198$ obj., held-out)} \\
\cmidrule(lr){2-5}\cmidrule(lr){6-9}
& \#Br. & $n_{f}$ & $n_{u}$ & Stab\%
& \#Br. & $n_{f}$ & $n_{u}$ & Stab\% \\
\midrule
Legolization   & 177.1 & 0.79 & 0.79 & 81 & 508.7 & 5.33 & 4.60 & 76.5 \\
BrickGPT       & 140.5 & \textbf{0} & \textbf{0} & \textbf{100} & 347.2 & 3.28 & 3.08 & 74.2 \\
STABLE         & 221.7 & 1.74 & 2.05 & 57 & 330.7 & 12.98 & 12.62 & 14.0 \\
\textbf{Ours}                 & \textbf{129.5} & \textbf{0} & 0.29 & \textbf{97.6} & 343.1 & \textbf{0} & \textbf{0} & \textbf{100} \\
\textbf{Ours (col.)}          & --- & --- & --- & --- & \textbf{340.1} & \textbf{0} & 0.45 & 98.0 \\
\midrule
\multicolumn{9}{l}{\emph{Ablation}} \\
\textsc{naive}       & 131.8 & 0.00 & 0.43 & 90.5 & 350.2 & 0.00 & 0.12 & 96.0 \\
\textsc{no-supp.}    & 130.5 & 0.00 & 0.07 & 97.6 & 348.2 & 0.00 & 0.02 & 98.0 \\
\textsc{no-look.}    & 130.3 & 0.00 & 0.29 & 97.6 & 355.0 & 0.00 & 0.00 & 100.0 \\
\textsc{no-repair}   & 127.5 & 2.17 & 2.05 & 52.4 & 339.3 & 0.00 & 0.46 & 98.0 \\
\bottomrule
\end{tabular*}
\caption{Building stage. Baselines vs.\ ours (top) and
component ablation (bottom) on \textsc{s2b} and held-out \textsc{omni};
Ours (col.) adds per-brick color. \#Br.: brick count;
$n_f$/$n_u$: floating/locally statically unbalanced bricks; Stab\%: fully stable models.
\textbf{Bold}: best among (near-)fully stable methods. Dashes: metric that remains the same 
reported by that method.}
\label{tab:stageB}
\end{table}

\section{Conclusion}
We presented \methodname{}, a complete pipeline that turns casual photographs into buildable brick models by coupling a resolution-conditioned occupancy-completion network with a buildability-aware greedy builder. Recasting coarse voxelization as budget allocation, the learned completion is perceptually the most faithful selector while remaining the most connected under a matched budget, and a single set of weights transfers across resolutions. The builder in turn yields assemblies with zero floating components at the lowest brick count on unconstrained held-out geometry, and the exported instructions are directly buildable by hand. Together these results indicate that how a discrete budget is spent bounds what any downstream assembly can achieve, so allocation and buildability are best co-designed rather than treated as independent stages. Several directions remain open: the intrinsic ceiling of low-resolution occupancy invites higher-resolution or adaptive budget allocation, and coupling brick-count minimization with stability more tightly could further improve the parsimony--stability balance.

\section*{Acknowledgments}
This work was partially supported by National Natural Science Foundation of China (Grant No. 62441619), Guangdong Basic and Applied Basic Research Foundation (Grant No. 2022A1515110411, Grant No. 2023A1515012883, and Grant No. 2024A1515240009), Shenzhen Science and Technology Program (Grant No. JCYJ20240813113604006, Grant No. KJZD20240903095730039), and Guangxi Science and Technology Planning Project (Grant No. AA23062031-2, Grant No. AA23062073-2).

\bibliography{aaai2027}

\clearpage
\appendix


\section{Overview of Supplementary Material}
\label{supp:overview}

This supplementary material provides implementation details and additional
analyses that support the main paper. We first give an expanded description of
the full pipeline (Sec.~\ref{supp:pipeline}) and of the data splits
(Sec.~\ref{supp:splits}), then implementation details for the voxelization
network and the stability surrogate (Sec.~\ref{supp:implementation}). We next
detail the evaluation protocol, metric definitions, and statistical testing
(Sec.~\ref{supp:evaluation}), followed by additional quantitative results
(Sec.~\ref{supp:additional_quantitative})---an architecture/capacity ablation,
a hyperparameter-selection sanity check, the fidelity--buildability trade-off, the full
perceptual-metric table, the geometry guardrail, and hyperparameter
settings. We close with an analysis of the scope of our buildability claims
(Sec.~\ref{supp:scope}), a justification of the lazy-greedy oracle construction
(Sec.~\ref{supp:celf}), a discussion of concurrent systems not run as
baselines (Sec.~\ref{supp:concurrent}), and limitations with future directions
(Sec.~\ref{supp:limitations}).

\section{Pipeline Details}
\label{supp:pipeline}

The full pipeline turns a handful of casual photographs into a colored,
hand-buildable brick model through four stages, each consuming the previous
stage's output.

\paragraph{(1)~Pose-free reconstruction.} From a few unposed input photographs,
a pose-free reconstructor (FreeSplatter~\cite{freesplatter} in our
implementation) predicts geometry and appearance without camera calibration;
we extract a watertight textured mesh via TSDF
fusion~\cite{curless1996volumetric} and Marching
Cubes~\cite{lorensen1987marching}. Output: a textured mesh with per-vertex
color.

\paragraph{(2)~Budgeted occupancy completion.} The mesh is voxelized on a coarse
lattice, and a single resolution-conditioned network selects, in one
feed-forward pass, which surface voxels to fill so as to best preserve
appearance under a target occupied-voxel budget $n_{\mathrm{occ}}(\rho)$
(main paper, voxelization stage). Output: a completed occupancy grid at the target
budget, with per-voxel color sampled from the mesh (Sec.~\ref{supp:palette}).

\paragraph{(3)~Buildable brick assembly.} Each layer's $4$-connected region is
partitioned into non-overlapping library bricks by a support- and
look-ahead-aware greedy placement, and a deterministic, provably terminating
repair grounds every floating component (main paper, building stage). Static
stability is certified post hoc by the exact force-balance solver, with the
distilled surrogate (Sec.~\ref{supp:surrogate}) used only as a fast in-loop
screen. Output: a grounded brick assembly.

\paragraph{(4)~Color assignment.} Each voxel's continuous color is quantized to
the fixed LEGO palette with an adaptive, edge-preserving reduction, and every
brick inherits a single representative color (Sec.~\ref{supp:color},
Sec.~\ref{supp:palette}). Output: the final colored, layer-by-layer
buildable brick model.

Some of the stage descriptions above necessarily restate the main paper so
that this supplement is readable on its own; the main paper remains the
authoritative source for the method.

\section{Data Splits}
\label{supp:splits}
We partition OmniObject3D~\cite{omniobject3d} into disjoint sets that are used
consistently throughout the paper and this supplement. The completion network
is trained on the $3070$-object \texttt{rho} training split (one instance per
category withheld). The main-paper evaluation---voxelization, building, and
full-pipeline---is reported on the $198$-object held-out validation set.
Building-stage hyperparameters are selected on a separate
\emph{sweep-validation} set (\texttt{sweepval3}, $35$ objects from $12$
categories) that never appears in the training split; a hard \texttt{assert}
in the split builder verifies \emph{zero instance and zero category overlap}
with training. Tuning building-stage weights on this set therefore never
touches the training data, and it is kept small (mean occupancy $671$ vs.\
$2066$ for the full set) so a single configuration runs in minutes. The
network architecture and checkpoint are fixed on training loss and early
stopping, not on any evaluation-set metric (main paper, voxelization stage).

\section{Implementation Details}
\label{supp:implementation}

\subsection{Voxelization Network: Architecture and Training}
\label{supp:stagea_impl}

The completion network (\texttt{OccCompletionNet}) is a fully-convolutional
3D U-Net~\cite{ronneberger2015unet,cicek20163dunet} with two down/up-sampling
stages and skip connections. The three channel widths follow a fixed
$1{:}2{:}4$ ratio $c_1,c_2,c_3=(\mathrm{base},2\,\mathrm{base},
4\,\mathrm{base})$; our default \emph{main-result} model uses
$\mathrm{base}{=}8$, i.e.\ $8{\to}16{\to}32$ channels ($0.10$M parameters).
At resolution $R{=}32$ the spatial grid contracts $32^3{\to}16^3{\to}8^3$
through the encoder and mirrors back in the decoder.

\begin{table}[t]
\centering
\small
\setlength{\tabcolsep}{5pt}
\begin{tabular}{lcc}
\toprule
Stage & Spatial & Channels \\
\midrule
enc1                       & $32^3$ & $\mathrm{in}\to 8$ \\
down1 $+$ enc2             & $16^3$ & $8\to 16$ \\
down2 $+$ bottleneck       & $8^3$  & $16\to 32$ \\
up2 $+$ dec2 (skip enc2)   & $16^3$ & $32\to16,\ \mathrm{cat}(16{+}16)\to16$ \\
up1 $+$ dec1 (skip enc1)   & $32^3$ & $16\to8,\ \mathrm{cat}(8{+}8)\to8$ \\
head                       & $32^3$ & $8\to 1$ ($1^3$ conv, zero-init) \\
\bottomrule
\end{tabular}
\caption{Layer structure of the completion U-Net at $\mathrm{base}{=}8$,
shown for input resolution $R{=}32$. Skip connections concatenate the
matching encoder features into the decoder.}
\label{supp:tab_arch}
\end{table}

Each \texttt{ConvBlock} is two $3^3$ \texttt{Conv3d} layers, each followed by
GroupNorm~\cite{wu2018group} with $\min(8,C)$ groups and a \texttt{GELU}
activation. Down-sampling uses a stride-$2$ convolution and up-sampling a
\texttt{ConvTranspose3d}; the $1^3$ output head is zero-initialized so the
network starts from the identity fill. The density--resolution condition
$[R/64,\rho]$ is mapped by a small MLP to per-channel FiLM
parameters~\cite{perez2018film} $(\gamma,\beta)$ that modulate every
\texttt{ConvBlock}, so a single set of weights serves $R\in\{24,32,48\}$ and
generalizes to unseen resolutions, as shown by the resolution-generalization
figure in the main paper.

Training uses the \texttt{rho} split ($3070$ train) on the Voxel-train L40S
GPU (Sec.~\ref{supp:infra}), with the Adam optimizer at a $10^{-3}$ learning
rate. Phase~1 distills
the buildability-aware completion oracle via a ranking loss on the
surface-band scores and trains for $80$ epochs. Phase~2 refines the network
against rendered fidelity under a straight-through
estimator~\cite{bengio2013estimating} whose temperature is annealed
geometrically from $\tau{=}2.0$ to $0.1$, with a budget term pinning the
filled count to $n_{\mathrm{occ}}(\rho)$, and trains for $12$ epochs.

\subsection{Stability Surrogate}
\label{supp:surrogate}
All stability numbers in the main paper use the exact force-balance
solver. For use inside training, we additionally distill this solver into
a millisecond message-passing surrogate that predicts per-brick
stability, reaching a per-brick AUC of $0.984$ and transferring across
voxel resolutions without fine-tuning. Because it is a differentiable
drop-in for the MILP gate, it can be used as a soft stability signal
during optimization while all reported evaluation remains on the exact
solver.

\subsubsection{Surrogate Architecture}
\label{supp:surrogate_arch}
The surrogate is an encode--process--decode message-passing network over
the assembly graph, following the MetaLayer formulation. Bricks are nodes
(plus a single virtual \emph{ground} node); edges connect a brick in layer
$z$ to a brick in layer $z{+}1$ whose XY footprints overlap (a vertical
contact), and ground-level bricks to the ground node. Contact edges are
directed and instantiated in both orientations (down$\to$up and
up$\to$down) so the network can encode which brick presses on which;
same-layer neighbors are \emph{not} connected, as lateral adjacency
provides no vertical support.

All features are deliberately \emph{dimensionless} (counts and ratios
only)---this is what lets a single set of weights transfer across voxel
resolutions without fine-tuning; in particular only the \emph{relative}
height $z/z_{\max}$ enters and absolute coordinates are never used.
Table~\ref{supp:tab_surrogate_feat} lists the $11$ node and $4$ edge
features.

\begin{table}[t]
\centering\small
\setlength{\tabcolsep}{5pt}
\begin{tabular}{ll}
\toprule
\multicolumn{2}{l}{\emph{Node features} ($11$-dim)} \\
\midrule
$h,\ w$ & footprint dimensions \\
$\min(h,w)$ & shorter footprint side \\
$h\,w$ & footprint area \\
$\mathbf{1}[1{\times}X]$ & is a $1{\times}X$ brick \\
$\mathrm{mass}/\mathrm{mass}_{1\times1}$ & mass relative to a unit brick \\
$z/z_{\max}$ & relative height in the model \\
$n_{\mathrm{bot}},\ n_{\mathrm{top}}$ & bottom / top contact counts \\
$\mathbf{1}[\text{ground brick}]$ & rests on the ground \\
$\mathbf{1}[\text{ground node}]$ & is the virtual ground node \\
\midrule
\multicolumn{2}{l}{\emph{Edge features} ($4$-dim)} \\
\midrule
$n_{\mathrm{contact}}$ & \# overlapping contact cells \\
$n_{\mathrm{contact}}/\min(A_A,A_B)$ & contact relative to smaller area \\
$\mathbf{1}[\text{up}]$ & edge orientation (up / down) \\
$\mathbf{1}[\text{ground}]$ & is a ground-contact edge \\
\bottomrule
\end{tabular}
\caption{Dimensionless node and edge features of the stability surrogate
($A_A,A_B$ are the two incident brick areas). Using only counts and
ratios makes the surrogate resolution-agnostic.}
\label{supp:tab_surrogate_feat}
\end{table}

The network follows the standard encode--process--decode template:
\begin{itemize}
\item \textbf{Encoder:} two MLPs lift node features $11{\to}128{\to}128$
  and edge features $4{\to}128{\to}128$.
\item \textbf{Processor:} $M{=}5$ MetaLayer blocks. Each block first
  updates every edge from its two endpoints and its own state, then updates
  every node from its current state and the mean of incoming edge messages,
  with residual connections on both node and edge states.
\item \textbf{Decoder:} a node MLP $128{\to}128{\to}2$ with a dual head
  emitting a per-brick instability logit and a stability margin.
\end{itemize}
The network has $0.628$M parameters. Being purely message-passing over
dimensionless features, it is resolution-agnostic (trained at $20^3$,
applied at $16$--$64^3$).

It is trained on StableLego's per-brick stability labels (from
force-balance analysis) with a per-brick loss combining a
class-balanced instability term and a margin-regression term,
\begin{equation}
\label{eq:surrogate_loss}
\mathcal{L}=\mathrm{BCE}\big(\hat{s},\,y_{\mathrm{dead}};\,w^{+}\big)
+\lambda\,\mathrm{Huber}(\hat m, m),
\end{equation}
where $\hat s$ is the predicted instability logit, $y_{\mathrm{dead}}$ the
ground-truth instability label, $w^{+}{=}12.8$ the positive class weight,
and the Huber margin term ($\lambda{=}1$, $\delta{=}0.1$) is applied to
stable bricks only. Optimization uses Adam
($\mathrm{lr}\,10^{-3}$, weight decay $10^{-4}$), with plateau LR
scheduling and early stopping on validation AUC (best checkpoint
reported). Splits are by object (all bricks of an object share a split) to
prevent leakage.


\subsubsection{Distillation Results}
\label{supp:surrogate_results}
The surrogate is evaluated on a held-out test set of $110$ objects
($880$ train / $110$ validation / $110$ test, using StableLego dataset, balanced by category).
Because the training positive-weighting inflates predicted instability
probabilities, the decision threshold is calibrated on validation by
maximizing object-level agreement (optimum $\approx0.95$); the default
$0.5$ threshold is \emph{not} used, and re-reading at the calibrated
threshold is a decoding fix, not a change of model
(Table~\ref{supp:tab_surrogate}). The per-brick AUC is threshold-
independent at $0.984$, matching the value quoted in the main paper. At
the calibrated threshold the object-level buildability gate agrees with
the exact solver on $99.1\%$ of test objects.

\begin{table}[t]
\centering\small
\setlength{\tabcolsep}{6pt}
\begin{tabular}{lcc}
\toprule
Metric & thr.\ $0.5$ (uncal.) & thr.\ $0.95$ (cal.) \\
\midrule
AUC            & $0.984$ & $0.984$ \\
Accuracy       & $0.925$ & $0.984$ \\
Precision      & $0.462$ & $0.940$ \\
Recall         & $0.958$ & $0.801$ \\
F1             & $0.623$ & $0.865$ \\
Object-level agreement & $0.718$ & $\mathbf{0.991}$ \\
\quad GT-buildable correct  & $0.380$ & $0.980$ \\
\quad GT-unbuildable correct & $1.000$ & $1.000$ \\
\bottomrule
\end{tabular}
\caption{Surrogate distillation fidelity on the $110$-object test set
(per-brick unless noted). The AUC is threshold-independent; the two
columns differ only in the decision threshold, calibrated on validation.}
\label{supp:tab_surrogate}
\end{table}

\paragraph{Cross-resolution transfer.}
Trained only at $20^3$, the surrogate applies unchanged at other
resolutions. On a $32^3$ assembly (an object of $608$ bricks) the
per-brick AUC is $0.9955$ and accuracy $0.987$, confirming that the
\emph{ranking} of bricks by stability transfers; the absolute decision
threshold does not, since the calibrated $0.95$ cut from $20^3$ is no
longer optimal on the far more imbalanced $32^3$ distribution (dead
bricks ${\sim}1.2\%$), where F1 drops to $0.50$ even as AUC rises. In
practice the threshold is re-calibrated per deployment resolution.

\paragraph{Limitation.}
The per-brick score correlation is modest (Spearman $0.44$) because the
regression target saturates at the margin ceiling for the many clearly-
stable bricks, giving a weak continuous ranking signal. This does not
affect the binary instability verdict that the gate relies on, but a
downstream use requiring a smooth stability gradient field would be
limited by it. Separately, on our own OmniObject3D sweep data the
surrogate's deviation from the exact solver is one-sided (it
\emph{over}-predicts instability, i.e.\ it is conservative), with
object-level agreement $96.2\%$; we therefore use it only for trend/
ordering during optimization and report all final numbers from the exact
solver.

\subsection{Additional Implementation Notes}
\label{supp:implementation_notes}

\paragraph{Color availability.}
Colors are drawn from the $91$ solid LEGO colors
(Sec.~\ref{supp:palette}), and every brick carries a single color. We do not
model a per-size color availability: inventory is treated as effectively
unconstrained, so numerically any brick size can take any color. The
``closest same-size color'' substitution mentioned in the main paper is a
\emph{manual} substitution performed only during our physical hand-builds when
a required brick size is unavailable in stock; it is a physical-build step, not
a pipeline step.

\subsubsection{Color-Fidelity Term: Full Expressions}
\label{supp:color}
The main paper introduces the per-brick color-fidelity term
$\Phi_{\mathrm{col}}$ as an ``importance--rarity salience'' that makes
brick seams prefer color edges over uniform regions. Here we give the
exact expressions used in our implementation.

\paragraph{Per-voxel color and salience.}
Each surface voxel $v$ carries a discrete LEGO color id $c(v)$ (an index
into the assembly palette; interior voxels are colorless and are excluded
from every sum below). We attach to each colored surface voxel a scalar
salience weight
\begin{equation}
\label{eq:colorweight}
w(v) \;=\; \mathrm{imp}(v)\;\cdot\;\mathrm{rar}\!\big(c(v)\big),
\end{equation}
the product of a \emph{local isolation} (importance) factor and a
\emph{global rarity} factor.

The isolation factor measures how texturally isolated a voxel is within
its $3{\times}3{\times}3$ neighborhood $\mathcal{N}(v)$ (the $26$ neighbors).
Letting $\mathcal{N}_{\mathrm{occ}}(v)$ be the occupied neighbors and
$\mathcal{N}_{\mathrm{same}}(v)\subseteq\mathcal{N}_{\mathrm{occ}}(v)$ those
sharing the voxel's own color id,
\begin{equation}
\label{eq:imp}
\begin{aligned}
\mathrm{imp}(v) &\;=\; 1 - \frac{|\mathcal{N}_{\mathrm{same}}(v)|}
{|\mathcal{N}_{\mathrm{occ}}(v)|},\\
\mathrm{imp}(v) &\!:=\!1 \ \text{ if } |\mathcal{N}_{\mathrm{occ}}(v)|{=}0 .
\end{aligned}
\end{equation}
A voxel surrounded by same-color neighbors (flat, uniform region) gets
$\mathrm{imp}\!\to\!0$; an isolated color speck (e.g.\ an eye dot, a baked
edge) approaches $1$.

The rarity factor rewards globally scarce colors. Let $f_c$ be the number
of colored surface voxels carrying color id $c$, and let
$N=\sum_{c} f_c$ be the total count of colored surface voxels (the sum
runs over colored ids only; colorless interior voxels do not contribute).
Our default (\texttt{inv}) uses the inverse frequency
\begin{equation}
\label{eq:rar}
\mathrm{rar}(c) \;=\; \frac{N}{f_c},
\end{equation}
with two alternatives we also support, $\mathrm{rar}_{\log}(c)=-\log(f_c/N)$
and $\mathrm{rar}_{\sqrt{}}(c)=\sqrt{N/f_c}$; the inverse form performed best
and is used in all reported results. Setting $w(v)\equiv 1$ recovers the
plain cell-count behavior.

\paragraph{Per-brick term.}
A physical brick $b$ carries a single color, so it can honor only one of
the color ids present in its footprint. We use the salience weights in two
related but distinct roles. \emph{(i)~Assigning} the brick's single
representative color is a majority-style vote weighted by the local
isolation factor $\mathrm{imp}(v)$ alone:
\begin{equation}
\label{eq:cbest}
c^\star(b) \;=\; \arg\max_{c}\;\sum_{v\in b,\, c(v)=c} \mathrm{imp}(v),
\end{equation}
so that isolated texture voxels dominate the choice of which color a
mixed-color brick should take. \emph{(ii)~Scoring} the brick during greedy
placement uses the full salience $w(v)=\mathrm{imp}(v)\,\mathrm{rar}(c(v))$,
rewarding the honored mass and penalizing the misrepresented mass:
\begin{equation}
\label{eq:phicol}
\begin{aligned}
\Phi_{\mathrm{col}}(b)
= w_{\mathrm{col}}\Big[
&\sum_{\substack{v\in b\\ c(v)=c^\star(b)}}\!\! w(v)\\
&-\!\!\sum_{\substack{v\in b\\ c(v)\neq c^\star(b),\ \text{colored}}}\!\! w(v)
\Big],
\end{aligned}
\end{equation}
where $w_{\mathrm{col}}$ is the color weight ($w_{\mathrm{col}}{=}20$ in all
experiments; $w_{\mathrm{col}}{=}0$ disables the term for backward
compatibility). Using $\mathrm{imp}$ alone for the color \emph{vote} and the
full $w$ for the placement \emph{score} reflects our implementation:
the representative color follows local texture isolation, while the
global rarity factor additionally steers where the greedy merge chooses to
cut seams. A single-color brick incurs no penalty and scores
proportionally to its (weighted) area, matching the plain area reward;
a brick that swallows a high-salience rare/isolated voxel of a different
color is heavily penalized, pushing the greedy selector to give such
voxels their own small brick and preserve the color. This is the term
that recovers fine chromatic detail (e.g.\ the teddy's eyes, the
biscuit's baked rim) that a purely area-driven merge would erase.


\subsubsection{Color Assignment and Palette}
\label{supp:palette}
Color enters the pipeline in two decoupled steps: (i)~sampling a
continuous color for every occupied voxel from the reconstructed textured
mesh, and (ii)~quantizing those colors to a fixed LEGO color set with an
adaptive, edge-preserving reduction. Both run as preprocessing, before
brick placement, so the building stage sees a discrete color id per voxel.

\paragraph{Barycentric color sampling from the mesh.}
Given the reconstructed mesh $\mathcal{M}$ with per-vertex colors, we color
each occupied voxel by its center point $q$. We query an
\texttt{o3d} raycasting scene for the closest surface point to $q$, which
returns the hit triangle $(v_0,v_1,v_2)$ and its barycentric coordinates
$(u,v)$. The voxel color is the barycentric interpolation of the three
vertex colors,
\begin{equation}
\label{eq:bary}
c(q) \;=\; w\,\mathbf{c}_0 + u\,\mathbf{c}_1 + v\,\mathbf{c}_2,
\qquad w = 1-u-v,
\end{equation}
rather than the nearest-vertex color, so that colors vary smoothly when a
voxel center projects into a triangle interior. This yields a continuous
sRGB color per surface voxel, stored directly in the color channels of the
occupancy grid.

\paragraph{Diversity-aware CIELAB quantization.}
The continuous colors are quantized to the set of $91$ solid LEGO colors.
All color distances use the full \mbox{CIEDE2000}
difference (with the $S_L,S_C,S_H,R_T$ terms)~\cite{sharma2005ciede2000}, not a plain
CIE76 $\Delta E^{*}_{ab}$ Euclidean distance, because CIEDE2000 better
matches perceptual color proximity for the saturated hues common in toy
objects. Quantization proceeds in three stages and, crucially, never fixes
a target number of colors $k$; the palette size is adaptive per object.

\emph{(1)~Nearest-color mapping.} Each colored surface voxel is converted
to CIELAB and assigned the LEGO color minimizing its CIEDE2000 distance
(independent per voxel; no clustering).

\emph{(2)~Adaptive $\Delta E$ merging.} We then repeatedly merge the two
closest \emph{used} colors while their CIEDE2000 distance is below a
threshold $\Delta E_{\min}{=}10$, folding the smaller-area color into the
larger. This softly reduces near-duplicate colors without a preset $k$.

\emph{(3)~Long-tail connectivity test.} Colors covering less than
$0.5\%$ of the colored surface voxels are treated as a long tail. For each
tail color we run a $3{\times}3{\times}3$ connected-component analysis: if
its largest connected component still covers at least $0.3\%$ of the
voxels, the color is \emph{kept} (it is a spatially coherent texture
feature, e.g.\ an eye or a printed marking); otherwise it is scattered
noise and is folded into its nearest retained color under CIEDE2000. This
is what preserves small but meaningful chromatic details while discarding
speckle.

At build time a brick inherits the representative color of
Eq.~\eqref{eq:cbest}. When the required brick size is unavailable in that
color, we substitute the closest same-size color (again by CIEDE2000),
which accounts for the minor color differences noted in the real-world
builds.

\section{Evaluation Protocol and Metrics}
\label{supp:evaluation}

\subsection{Metrics and Evaluation Setup}
\label{supp:metrics_setup}

\paragraph{Baseline protocols.}
All voxelization-stage comparisons use the matched occupancy budget
$n_{\mathrm{occ}}(\rho)$. The compared selectors are derived from the same
frozen SDF, camera poses, and mesh, and are evaluated over
$\rho\in\{0.4,0.62,0.75\}$ and $R\in\{24,32,48\}$ unless otherwise stated.
The non-oracle voxelization baselines are SDF shell, Top-$K$, dilation,
6-separating, and 26-separating selectors; each is defined exactly, together
with the shared budget-matching rule, in Sec.~\ref{supp:selector_defs}. TSDF
fusion is reported as an oracle-input reference because it additionally uses
ground-truth depth, and is therefore excluded when marking the best non-oracle
method. The full pipeline
is compared with publicly available brick-construction systems,
BrickGPT~\cite{pun2025}, Legolization~\cite{luo2015}, and BrickLink
Studio~\cite{bricklinkstudio}, on the held-out OmniObject3D set at $R{=}24$.
For the building-stage comparison, BrickGPT, Legolization, and
STABLE~\cite{stable2026} are run on the same input grids over both
StableText2Brick and the held-out OmniObject3D subset.

\subsubsection{Voxelization Selectors: Exact Definitions}
\label{supp:selector_defs}
Every voxelization selector reads the \emph{same} frozen per-object cache: a
continuous signed-distance field $\phi$ evaluated at voxel centers on the $R^3$
lattice of the unit cube $[0,1]^3$ (so $\phi$ is in cube-normalized units and
the voxel edge is $h{=}1/R$), the same camera poses, and the same textured
mesh. No selector re-runs alignment, surface sampling, or reconstruction, so
fairness here is by construction rather than by statistical control. Two
derived sets are fixed once per object: the \emph{solid core}
$\mathcal{B}=\{v:\phi(v)\le 0\}$, the voxels whose centers lie inside the
surface, and the \emph{surface band}
$\mathcal{S}=\{v:\mathrm{cov}(v)>0\}\setminus\mathcal{B}$, where
$\mathrm{cov}(v)$ is the fraction of voxel $v$ covered by the mesh, so
$\mathcal{S}$ is exactly the shell of voxels the surface passes through but
whose centers fall outside it. The matched budget is
\begin{equation}
\label{eq:nocc}
n_{\mathrm{occ}}(\rho)\;=\;|\mathcal{B}|\;+\;\mathrm{round}\!\big(|\mathcal{S}|\,\rho\big),
\end{equation}
i.e.\ the core is always kept and only the band is contested. The selectors
are:

\begin{itemize}
\item \textbf{SDF shell} is the raw solid core $\mathcal{B}$: fill every voxel
  whose center is inside the surface, and nothing else. This is the default
  ``just voxelize the mesh'' behavior and is the one selector with \emph{no}
  budget parameter---at low $\rho$ it is denser than $n_{\mathrm{occ}}$ and at
  high $\rho$ sparser---so it is reported as an unbudgeted reference rather
  than as a same-budget competitor.
\item \textbf{Top-$K$} keeps the $n_{\mathrm{occ}}$ voxels of smallest $\phi$,
  a single global partial sort on the signed distance. Equivalently, it sweeps
  one global iso-level until the budget is exactly met. This is the strongest
  purely geometric selector in our comparison.(In other sections of this supplement, this same selector may be labeled as \emph{uniform}: ``uniform'' and ``Top-$K$'' denote the identical construction and differ only in naming.)
\item \textbf{Dilate} starts from $\mathcal{B}$ and applies rounds of
  $6$-connected binary dilation (the $3^3$ cross structuring element) until the
  count first reaches $n_{\mathrm{occ}}$, then trims back to the budget. It
  differs from Top-$K$ in that growth is forced to stay contiguous with the
  core instead of being free to open new components at a distance.
\item \textbf{6-separating} takes the touched set
  $\{v:|\phi(v)|\le \tfrac{1}{2}h\}$: voxels whose \emph{inscribed} sphere
  (radius $h/2$) reaches the surface. This is the thin, $6$-separating
  criterion of \texttt{binvox}-style thin voxelization.
\item \textbf{26-separating} takes
  $\{v:|\phi(v)|\le \tfrac{\sqrt{3}}{2}h\}$: voxels whose \emph{circumscribed}
  sphere (radius $\sqrt{3}h/2$, the half body diagonal) reaches the surface---%
  the conservative rasterization criterion.
\item \textbf{TSDF} fuses six depth observations of the \emph{ground-truth}
  mesh: for each of the six evaluation camera positions we cast a ray from the
  eye directly to every voxel center and take $\mathrm{sdf}_{\mathrm{view}} =
  t_{\mathrm{hit}} - d_{\mathrm{vox}}$, truncated to $\pm\mu$ with $\mu{=}2h$.
  Casting to voxel centers rather than rasterizing a depth image avoids image
  discretization, so this is equivalent to an infinite-resolution depth map
  under the same camera geometry as the rendered evaluation. Only observations
  with $\mathrm{sdf}_{\mathrm{view}}>-\mu$ vote (deeply occluded views abstain);
  a voxel with no valid observation is treated as interior ($-\mu$). A voxel is
  occupied when its fused value is ${\le}0$. Because TSDF consumes ground-truth
  depth it is an oracle-input reference, not a competing selector.
\item \textbf{Ours} keeps $\mathcal{B}$ and hardens the top
  $n_{\mathrm{occ}}-|\mathcal{B}|$ band voxels by predicted score, so the
  contested set is exactly the band and the realized count is
  $n_{\mathrm{occ}}$ by construction.
\end{itemize}

\paragraph{Budget matching.}
Top-$K$ and our network meet Eq.~\eqref{eq:nocc} by construction. The two
separating selectors and TSDF are threshold-defined, so their raw output does
not land on the budget exactly; they are matched by a \emph{single shared}
routine, so that no selector gains or loses from the matching rule itself.
Given a raw occupancy set and two per-voxel keys---a \emph{trim} key and a
\emph{fill} key---the routine is:
\begin{itemize}
\item if the raw set is \emph{larger} than $n_{\mathrm{occ}}$, keep the
  $n_{\mathrm{occ}}$ occupied voxels of smallest trim key, so the retained
  voxels are those hugging the surface most tightly;
\item if it is \emph{smaller}, add the unoccupied voxels of smallest fill key,
  so the additions are the most interior / nearest-surface voxels, until the
  budget is met.
\end{itemize}
The trim key is $|\phi|$ for the separating selectors and $|\mathrm{tsdf}|$ for
TSDF; the fill key is the corresponding \emph{signed} value ($\phi$, resp.\ the
fused value), which is what makes ``smallest'' mean most-interior on the fill
side and closest-to-surface on the trim side. Dilate uses the same routine with
$\phi$ as both keys. Selection is a deterministic partial sort, and the
realized voxel count is asserted equal to $\min(n_{\mathrm{occ}},R^3)$ for
every selector and object.

\paragraph{Caliber note on the separating selectors.}
The two separating selectors are implemented as a distance-field test at voxel
centers rather than as per-triangle separating-axis (SAT) tests. The two
thresholds bracket exact SAT: the inscribed-sphere set is contained in the
SAT-touched set, which is in turn contained in the circumscribed-sphere set, so
$6$-sep is never more permissive and $26$-sep never more restrictive than SAT.
The sets differ only for voxels grazed at a corner. Because the identical field
and thresholds are used for every object and resolution, this affects neither
the comparison nor the reported ordering.

\paragraph{Rendered-view metrics.}
Perceptual fidelity is measured on rendered views of the predicted output and
the target. The headline perceptual metrics are MS-SSIM~\cite{wang2003msssim}
and LPIPS~\cite{zhang2018lpips} with an AlexNet backbone
(LPIPS$_{\mathrm{a}}$). The full rendered-image metric set additionally
includes PSNR, single-scale SSIM~\cite{wang2004ssim},
DISTS~\cite{ding2020dists}, VGG-LPIPS, and normal consistency. PSNR, SSIM,
MS-SSIM, DISTS, and LPIPS are computed on color renderings; normal consistency
measures cosine agreement between rendered normal maps. Metrics are averaged
over the rendered views and then aggregated over the object set, with
additional averaging over $\rho$ and $R$ where indicated by the table caption.
For the oracle construction used in Stage~A training, the rendering loss is
computed from six-view renderings that aggregate depth, silhouette, normal,
and LPIPS terms.

\paragraph{Geometry metrics.}
Geometry is used as a guardrail rather than as the primary objective. Before
geometry evaluation, each object is normalized to the unit cube, and both the
predicted and target surfaces are sampled into dense point sets. Chamfer
distance is the symmetric mean nearest-neighbor distance between the predicted
and target surface point sets under this unit-cube normalization. The F-score
at a distance threshold $\tau$ is the harmonic mean of precision (fraction of
predicted points within $\tau$ of the target surface) and recall (fraction of
target points within $\tau$ of the predicted surface); we report it at both
$\tau{=}0.02$ and the tighter $\tau{=}0.01$, in unit-cube coordinates. Normal
consistency (NC) is the mean absolute cosine similarity between the surface
normal at each predicted point and the normal at its nearest target point (and
symmetrically), so that $\mathrm{NC}{=}1$ denotes perfectly aligned surface
orientation; it measures agreement of local surface orientation rather than
point position.

\paragraph{Buildability metrics.}
Voxelization buildability is measured directly on the completed occupancy
grid. We report the number of connected components (\#Comp), the fraction of
floating voxels, the fraction of unsupported voxels, and, when repair is
applied, whether the repair restores a single grounded component. Building
stage metrics are computed on the final brick assembly: \#Br.\ is the brick
count, $n_f$ is the number of floating bricks, $n_u$ is the number of
force-balance-unstable bricks under the exact solver, and Stab\% is the
fraction of models with zero floating and zero unstable bricks.

\subsection{Exact Force-Balance Solver}
\label{supp:solver}
All reported stability numbers are certified by an exact force-balance
mixed-integer program, StableLego's global force-balance
MILP~\cite{stablelego}. We state the formulation by name rather than
re-deriving it. Each brick carries per-contact unknowns---a normal force and a
friction/clutch force at the convex stud contacts; the constraints impose
per-brick force and moment balance, with equal-and-opposite contact forces
applied pairwise between touching bricks; the objective minimizes the total
force residual plus a small $\alpha,\beta$ regularization. The physical
constants are a per-stud clutch capacity $T{=}0.98$\,N, a per-cell mass
$2.87{\times}10^{-4}$\,kg, a unit footprint of $7.8\,\text{mm}\times
9.6\,\text{mm}$, and an equality tolerance $\texttt{EQ\_TOL}{=}10^{-6}$; the
MILP is solved with Gurobi under \texttt{MIPFocus}${=}1$.

\paragraph{Alignment with the main-paper $n_u$.}
The unstable-brick count $n_u$ is exactly the main-paper quantity: a brick is
counted when its force residual exceeds the tolerance, and $n_u$ sums such
bricks. The floating count $n_f$ is a \emph{separate}, brick-level union--find
count of components not connected to the ground (two bricks are connected when
their footprints overlap across adjacent layers, with the ground at $z{=}0$).
The stability rate is $\mathrm{Stab\%}$, the fraction of models with $n_f{=}0$
and $n_u{=}0$. The qualifier ``local'' in \emph{locally unbalanced} refers to
the fact that the per-brick reading treats a neighbor's support as given rather
than propagating instability transitively---precisely the notion the
message-passing surrogate (Sec.~\ref{supp:surrogate}) is distilled to
reproduce.

\paragraph{Time limit, timeout, and re-run rule.}
The per-assembly time limit is $300$\,s ($120$\,s for the surrogate-screened
sweep stage). A timeout is \emph{detected} as $n_u{=}n_{\mathrm{brick}}$ with
solver status $\neq$ \texttt{OPTIMAL}: the solver returns ``all bricks dead''
as a degenerate artifact rather than a genuine verdict. Such assemblies are
re-attempted once at a $1800$\,s limit, applied identically to all four
methods; only assemblies still unsolved at $1800$\,s are excluded, and they are
\emph{never} counted as unstable. Under this rule the recovery counts are
$27/27$ for our method, $18/18$ for BrickGPT, and $34/34$ for Legolization, so
zero assemblies are excluded for any method. This is empirical evidence, not a
guarantee; we note that at $R{=}32$ some assemblies remain unsolved even at
$1800$\,s, which is why, for a fair comparison, we do not use $R{=}32$ or finer
resolutions.

\subsection{Baseline System Settings}
\label{supp:baseline_settings}
We detail how each comparison system is run, in addition to the matched-budget
protocol stated in the main paper.

\paragraph{Full-pipeline baselines.}
All full-pipeline baselines consume the same input mesh as our method.
BrickGPT~\cite{pun2025} is driven through the \texttt{mesh2brick} module
shipped in its public repository, which uses Open3D for voxelization.
Legolization~\cite{luo2015} is run on the same Open3D-derived input as
BrickGPT, because the public Maya-plugin implementation is not functional;
using an identical front end keeps the comparison controlled. BrickLink
Studio~\cite{bricklinkstudio} is run through its built-in ``import 3D mesh /
convert to brick'' function.

\paragraph{Building-stage baselines.}
For the building-stage comparison, all methods are given the identical input
grid: the SDF-uniform voxelization dilated by $0.2$. BrickGPT is natively a
$20^3$ method, but because it is not a learned generator but an exact
Gurobi solve, it can be manually set to $R{=}24$; enlarging the grid only
increases solve time and does not change the result, so the comparison stays
faithful. STABLE~\cite{stable2026} is a trained-grid method, but the approach
itself admits enlargement to $R{=}24$. For a rigorous comparison we therefore
report STABLE on both its native benchmark (StableText2Brick) and the enlarged
OmniObject3D subset. To confirm that this enlargement does not penalize
STABLE's stability under our exact-solver protocol, we run it at its native $R{=}20$ and at the generalized
$R{=}24$ on the sweepval3, both certified with Gurobi
under the identical graph construction as our exact stability check (footprint
on XY, $z$ the layer height). Enlargement does not degrade STABLE
(Table~\ref{supp:tab_stable_r24}): the unstable-brick fraction and count
are essentially unchanged (indeed slightly lower at $R{=}24$), so reporting
STABLE at $R{=}24$ is fair.

\begin{table}[t]
\centering\small
\setlength{\tabcolsep}{8pt}
\begin{tabular}{lcc}
\toprule
& $R{=}20$ (native) & $R{=}24$ (generalized) \\
\midrule
\#Brick      & $181.7$ & $237.0$ \\
$n_u$ fraction & $10.30\%$ & $7.72\%$ \\
$n_{u}$  & $19.00$ & $18.00$ \\
$n_{f}$ & $19.00$ & $20.33$ \\
all-stable   & $3/35$ & $2/35$ \\
\bottomrule
\end{tabular}
\caption{STABLE at its native $R{=}20$ vs.\ the generalized $R{=}24$ on the
$35$ sweep-validation objects, both certified with Gurobi under the same graph
construction as our exact stability check (footprint on XY, $z$ the layer
height). Enlarging to $R{=}24$ leaves STABLE's stability essentially unchanged,
confirming the $R{=}24$ report is not disadvantageous to it.}
\label{supp:tab_stable_r24}
\end{table}

\subsection{Statistical Testing}
\label{supp:stats}

All significance stars in the main paper (the voxelization-stage tables)
denote a two-sided paired Wilcoxon signed-rank test of each baseline against
\emph{Ours (net)}, computed on per-object metric differences. Pairing is per
object over the $198$ held-out validation objects, pooled across the reported
occupancy levels $\rho\in\{0.4,0.62,0.75\}$ and resolutions
$R\in\{24,32,48\}$. We use the Wilcoxon signed-rank test rather than a paired
$t$-test because the per-object metric distributions are non-normal and
contain outliers; the rank-based test makes no normality assumption. Star
levels are *$p{<}0.05$, **$p{<}0.01$, ***$p{<}0.001$; entries without a star
are not statistically distinguishable from \emph{Ours} at the $0.05$ level
(e.g., normal consistency vs.\ the budgeted top-$K$/dilation selectors, and
F-score vs.\ the 6-separating voxelizer). Because most differences are extreme
($p$ well below $10^{-10}$), we report star levels only rather than exact
$p$-values; the qualitative conclusions are unaffected.



\subsection{Random Seeds and Determinism}
\label{supp:seeds}
The building stage is deterministic once hashing is fixed: the only source
of run-to-run variation is the iteration order of Python \texttt{set}s in
the greedy merge, which affects tie-breaking between equally scored
candidates. We therefore fix \texttt{PYTHONHASHSEED=0} across the entire
pipeline (the entry scripts re-exec themselves with this variable set so
it takes effect before interpreter start-up). With this fixed, the
pipeline is bit-for-bit reproducible: re-running the full chain reproduces
identical per-object brick counts, and the
preprocessing reproduces the cached occupancy grids and color codes
exactly. The completion network is trained with fixed seeds for the data
split, weight initialization, and data loader; the split is additionally
frozen on disk so that the train/validation/test partition is identical
across all runs.

\section{Computing Infrastructure}
\label{supp:infra}
Our experiments run on two AWS GPU instances that share an identical
 conda environment.

\paragraph{Machines.}
The \emph{Voxel-train} machine is a \texttt{g6e.4xlarge}
(\texttt{ap-northeast-1a}): one NVIDIA L40S GPU ($48$\,GB GDDR6 ECC, Ada
Lovelace \texttt{sm\_89}), $16$ vCPUs (3rd-gen AMD EPYC), and $128$\,GiB RAM,
on the AWS Deep Learning AMI (Ubuntu $22.04$) with NVIDIA driver $570.133.20$
and system CUDA $12.4$. The \emph{Testing} machine is a \texttt{g5.4xlarge}:
one NVIDIA A10G GPU ($24$\,GB), $16$ vCPUs, and $62$\,GB RAM, with driver
$580.126.09$. Both use the same software stack: Python $3.9$, PyTorch
$2.4.1$ (CUDA $12.1$ wheels; the L40S host CUDA is $12.4$, the PyTorch wheels
remain \texttt{cu121}), NumPy $2.0.2$, Open3D $0.19.0$ (mesh handling and
barycentric color sampling, EGL surfaceless), trimesh $4.11.2$, and
scikit-image $0.24.0$. Six/twelve-view renderings for the perceptual objective
and the final evaluation are produced by a headless EGL rasterizer; all
perceptual metrics (PSNR, SSIM, MS-SSIM, LPIPS $0.1.4$, DISTS) are computed
with \texttt{piq} $0.8.0$.

\paragraph{What runs where.}
The voxelization stage---all preprocessing (saliency back-projection, SDF and
surface-band caching, oracle target generation), the resolution-conditioned
completion training (Phases~1 and~2), the architecture/capacity ablation, and
the stability-surrogate distillation---runs on the Voxel-train L40S. The
compact main-result network ($\mathrm{base}{=}8$) is retrained, and the
perceptual and geometry evaluations are run, on the Testing A10G; both machines
use the identical environment, so the split does not affect results. Exact
static stability is certified with the Gurobi force-balance MILP solver
(Sec.~\ref{supp:solver}); the distilled stability surrogate runs in PyTorch on
CPU at ${\sim}0.02$\,s per object.

\paragraph{Renderer substitution on Ada GPUs.}
On the L40S (\texttt{sm\_89}, CUDA $12.4$) the forward renderer of the
\texttt{diff\_gaussian\_rasterization} kernel that FreeSplatter depends on (the
\texttt{ashawkey} fork) fails deterministically---it reports a fixed, absurd
memory figure, and recompiling for \texttt{sm\_89} does not fix it. We therefore
replace the \texttt{render()} call in \texttt{gaussian\_utils.py} with
\texttt{gsplat} $1.5.3$ rasterization (\texttt{render\_mode="RGB+ED"},
forward-only; no differentiability is needed here). Reproducers on Ada-class
cards will hit the same failure, so we flag the substitution explicitly.
Building the native kernels additionally requires \texttt{CC=gcc-11},
\texttt{CXX=g++-11} (the environment's default GCC $15$ exceeds the \texttt{nvcc}
ceiling), and running requires \texttt{EGL\_PLATFORM=surfaceless}. Two
dependencies are not \texttt{pip}-installable directly and must be built:
\texttt{diff\_gaussian\_rasterization} (pinned to a fixed commit) and
\texttt{torchmcubes} (compiled from source).

\paragraph{Timing.}
The two GPUs differ by roughly an order of magnitude on the oracle workload: the
same oracle task (\texttt{alex-fp16-r16} at $R{=}32$) takes ${\sim}60$\,s per
object on the A10G versus ${\sim}7$\,s per object on the L40S.

\section{Additional Quantitative Results}
\label{supp:additional_quantitative}

\subsection{Architecture and Capacity Ablation}
\label{supp:stagea_ablation}

We first ablate the completion network's architecture and capacity
(Table~\ref{supp:tab_ablation}), then evaluate the chosen default model on
the \texttt{sweepval3} set (Table~\ref{supp:tab_indeptest}).

Table~\ref{supp:tab_ablation} reports the full ablation summarized in the
main paper. All variants share the training protocol above and differ in
a single design choice; we report validation IoU (against the completion
oracle) at $R\in\{24,32,48\}$ and $\rho{=}0.62$, with the two extreme fill
levels $\rho\in\{0.4,0.75\}$ for reference. Three conclusions hold. (i)~The
conditioning-injection scheme is not load-bearing: FiLM, channel
concatenation, and bottleneck cross-attention---all on the same 3D U-Net
backbone---agree to within $0.006$ IoU, and full-layer FiLM is
unnecessary (bottleneck-only matches it). (ii)~Group normalization is
essential: removing it makes training \emph{diverge}, with validation IoU
frozen at its initial $0.620$; skip connections
contribute a smaller but real $-0.010$. (iii)~Capacity is not the
bottleneck: scaling $\mathrm{base}$ from $8$ ($0.10$M) to $64$ ($5.68$M),
a $57{\times}$ increase, moves IoU only within noise ($0.653$--$0.657$),
which is why we adopt the smallest $\mathrm{base}{=}8$ model as the default.

\begin{table*}[t]
\centering\small
\setlength{\tabcolsep}{4pt}
\begin{tabular}{llcccc}
\toprule
Variant & Group & Params & IoU ($R24/32/48$) & $\rho0.4$ & $\rho0.75$ \\
\midrule
FiLM (full, ref.)      & inject.\ & $1.45$M & $0.695/0.653/0.623$ & $0.449$ & $0.774$ \\
concat 2ch             & inject.\ & $1.40$M & $0.691/0.653/0.622$ & $0.440$ & $0.770$ \\
cross-attn @bottleneck & inject.\ & $1.49$M & $0.688/0.648/0.618$ & $0.441$ & $0.770$ \\
FiLM @bottleneck only  & location & $1.42$M & $0.695/0.654/0.621$ & $0.444$ & $0.772$ \\
no GroupNorm           & norm     & $1.45$M & \multicolumn{1}{c}{\emph{fails to train}} & --- & --- \\
no skip                & struct.\ & $1.31$M & $0.675/0.647/0.620$ & $0.433$ & $0.768$ \\
\midrule
base$=8$ (default)     & capacity & $\mathbf{0.10}$M & $0.653$ (avg)$^{\S}$ & $0.440$ & $0.770$ \\
base$=16$              & capacity & $0.38$M & $0.654$ (avg) & $0.442$ & $0.770$ \\
base$=24$              & capacity & $0.83$M & $0.656$ (avg) & $0.441$ & $0.772$ \\
base$=32$              & capacity & $1.45$M & $0.657$ (avg) & $0.448$ & $0.774$ \\
base$=48$              & capacity & $3.22$M & $0.657$ (avg) & $0.448$ & $0.773$ \\
base$=64$              & capacity & $5.68$M & $0.656$ (avg) & $0.441$ & $0.773$ \\
\bottomrule
\end{tabular}
\caption{Architecture and capacity ablation on the \texttt{rho} split
(same split as the main results). Validation IoU vs.\ the completion
oracle; ``$R24/32/48$'' column is at $\rho{=}0.62$. All runs are $80$
epochs. \emph{no-norm} does not train (IoU frozen at initialization).
$^{\S}$The default model is within noise of the largest ($5.68$M) model
despite using $57{\times}$ fewer parameters.}
\label{supp:tab_ablation}
\end{table*}

As a hyperparameter-selection sanity check,
Table~\ref{supp:tab_indeptest} reports the \texttt{base}${=}8$
network on the \texttt{sweepval3} set, aggregated across $R\in\{24,32,48\}$ and
$\rho\in\{0.4,0.62,0.75\}$ ($9$ cells), against the two same-budget
geometric selectors (an SDF baseline and uniform voxelization). This
check confirms that our selection set touches neither the training nor
the main-evaluation data and that the network still leads on it; because
\texttt{sweepval3} is deliberately smaller and easier (mean occupancy
$671$ vs.\ $2066$ for the held-out set), its absolute numbers are
\emph{not comparable} to the main-paper $198$-object results and are
reported only for this sanity-check purpose.

\begin{table}[t]
\centering\small
\setlength{\tabcolsep}{4pt}
\begin{tabular}{lccc}
\toprule
Metric & net (\texttt{base}${=}8$) & SDF base & uniform \\
\midrule
PSNR $\uparrow$        & $\mathbf{25.974}$ & $25.417$ & $24.866$ \\
SSIM $\uparrow$        & $\mathbf{0.9430}$ & $0.9388$ & $0.9342$ \\
MS-SSIM $\uparrow$     & $\mathbf{0.9574}$ & $0.9456$ & $0.9440$ \\
LPIPS$_{\mathrm{vgg}}\downarrow$ & $\mathbf{0.0793}$ & $0.0842$ & $0.0836$ \\
LPIPS$_{\mathrm{alex}}\downarrow$ & $\mathbf{0.0735}$ & $0.0859$ & $0.0786$ \\
DISTS $\downarrow$     & $\mathbf{0.1881}$ & $0.1980$ & $0.1905$ \\
\midrule
floating $\downarrow$   & $\mathbf{0.0000}$ & $0.0108$ & $\mathbf{0.0000}$ \\
n\_comp $\downarrow$    & $1.035$ & $1.324$ & $\mathbf{1.003}$ \\
ungrounded $\downarrow$ & $\mathbf{0.0000}$ & $0.0024$ & $\mathbf{0.0000}$ \\
unsupported $\downarrow$& $0.0290$ & $0.0317$ & $\mathbf{0.0272}$ \\
\bottomrule
\end{tabular}
\caption{Hyperparameter-selection sanity check: retrained
\texttt{base}${=}8$ network on the \texttt{sweepval3} set, aggregated over
$R\in\{24,32,48\}\times
\rho\in\{0.4,0.62,0.75\}$. 
This set is deliberately smaller and easier (mean occupancy $671$ vs.\
$2066$ for the held-out set) and touches neither the training nor the
main-evaluation data; its absolute values are therefore \emph{not
comparable} to the main-paper $198$-object results and serve only to verify
that hyperparameter selection does not touch train/eval data and that the
network still leads.}
\label{supp:tab_indeptest}
\end{table}

\subsection{Fidelity--Buildability Trade-off Across Selectors}
\label{supp:pareto}
\begin{figure}[t]
\centering
\includegraphics[width=\columnwidth]{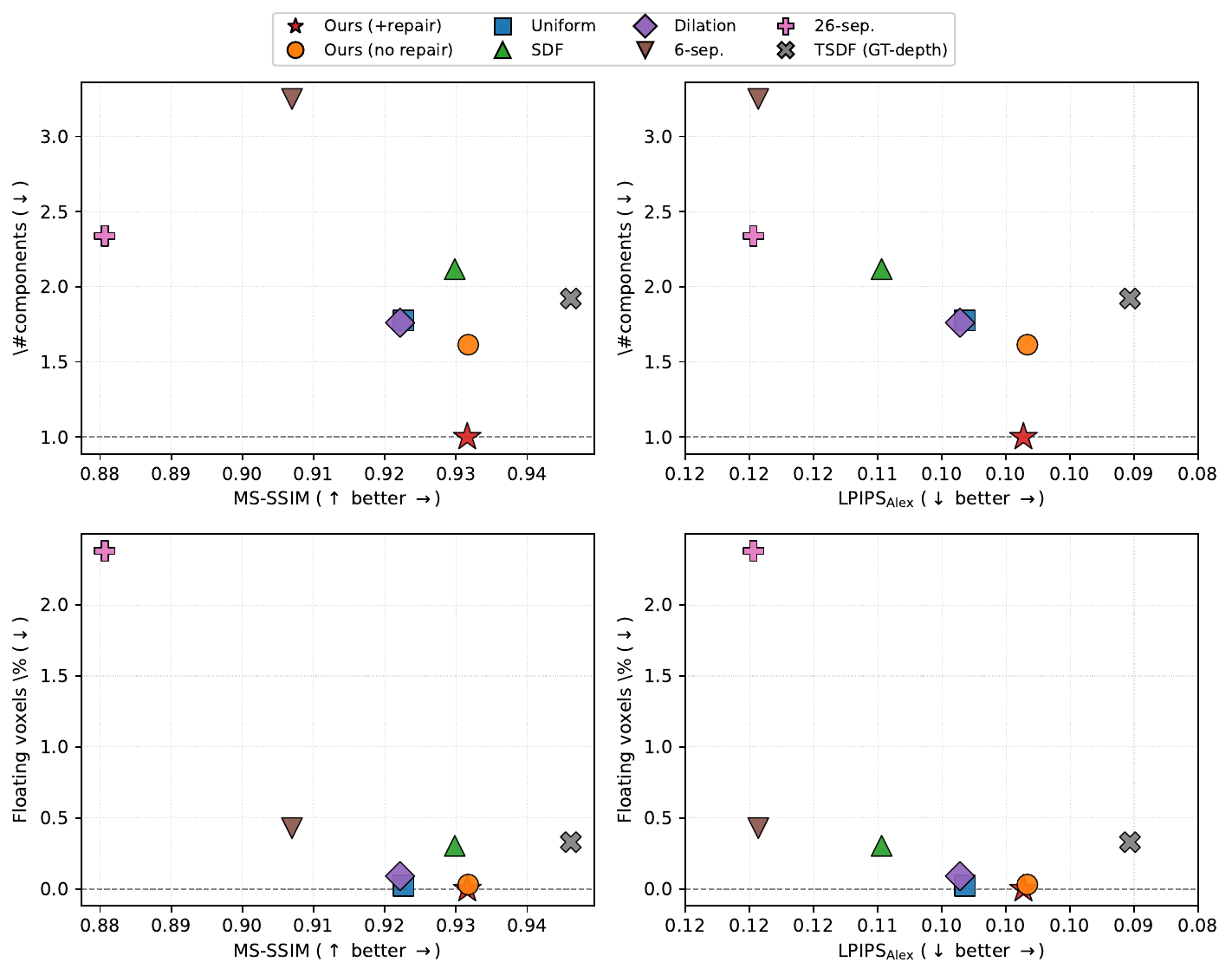}
\caption{Fidelity vs.\ buildability for all Stage-A voxelization
selectors, aggregated over $R\in\{24,32,48\}\times\rho\in\{0.4,0.62,0.75\}$
on the $198$-object held-out set (\texttt{base}${=}8$). Each axis is oriented so that
\emph{right} is better in fidelity and \emph{lower} is better in
buildability; the ideal region is the bottom-right corner. \textbf{Top:}
buildability as \#connected components ($\downarrow$, dashed line at the
ideal value $1$). \textbf{Bottom:} buildability as floating-voxel fraction
($\downarrow$). \textbf{Left/right:} fidelity as MS-SSIM / LPIPS$_{\mathrm{Alex}}$
(LPIPS axis reversed, true values on ticks). Ours ($+$repair) uniquely
occupies the ideal corner; TSDF uses ground-truth depth and is an
appearance upper bound that is not buildable.}
\label{supp:fig_pareto}
\end{figure}

Figure~\ref{supp:fig_pareto} places every voxelization selector in the
fidelity--buildability plane, aggregated over all nine
$R\times\rho$ cells on the $198$-object held-out set. Our full method (network $+$
grounding repair) is the only selector that reaches the ideal corner:
best-in-class perceptual fidelity among the constructive selectors while
attaining a single connected component (\emph{n\_comp}${=}1$) with zero
floating voxels. Two comparisons are instructive. First, grounding repair
moves the network from \emph{n\_comp}${\approx}1.6$ to exactly $1$ and
floating to $0$ \emph{without measurably changing fidelity} (MS-SSIM
$0.9317\!\to\!0.9316$, negligible loss), confirming that connectivity is restored at no
appearance cost. Second, the TSDF selector---fed \emph{ground-truth}
depth, hence an appearance upper bound---attains the highest fidelity but
remains unbuildable (\emph{n\_comp}${\approx}1.9$, nonzero floating),
showing that buildability must be targeted during training rather than
inherited from accurate geometry.

\subsection{Full Perceptual Metrics}
\label{supp:fullmetrics}
The main-paper end-to-end table (headline MS-SSIM and LPIPS$_{\mathrm{a}}$)
reports only two perceptual metrics for space. Table~\ref{supp:tab_e2e_full}
gives the complete six-metric set on the same held-out OmniObject3D set
($198$ objects, $R{=}24$), rendered from $12$ views at $256^2$ against the
target. The two headline columns match the main paper exactly; the
additional PSNR, SSIM, VGG-LPIPS, and DISTS all agree with the headline
conclusion---ResemBrick is best on every perceptual metric.

\begin{table*}[t]
\centering\small
\setlength{\tabcolsep}{4pt}
\begin{tabular}{lcccccc}
\toprule
Method & PSNR$\uparrow$ & SSIM$\uparrow$ & MS-SSIM$\uparrow$
& LPIPS$_{\mathrm{a}}\downarrow$ & LPIPS$_{\mathrm{v}}\downarrow$
& DISTS$\downarrow$ \\
\midrule
BrickGPT      & 24.30 & 0.940 & 0.940 & 0.081 & 0.097 & 0.233 \\
Legolization  & 24.33 & 0.939 & 0.940 & 0.082 & 0.098 & 0.235 \\
BrickLink Studio & 16.88 & 0.879 & 0.733 & 0.204 & 0.142 & 0.293 \\
\midrule
\textbf{Ours} & \textbf{26.05} & \textbf{0.949} & \textbf{0.955}
& \textbf{0.075} & \textbf{0.089} & \textbf{0.228} \\
\bottomrule
\end{tabular}
\caption{Full-pipeline perceptual metrics on the held-out OmniObject3D
set ($198$ objects, $R{=}24$; $12$ views at $256^2$). Best in
\textbf{bold}. BrickLink Studio is a manual-authoring commercial converter
reported at lower numeric precision. Ours wins all six metrics.}
\label{supp:tab_e2e_full}
\end{table*}

\subsection{Geometry Guardrail (Stage-A)}
\label{supp:geom}
Geometry is a \emph{guardrail}, not the objective: our claims are perceptual,
and geometric fidelity only has to remain competitive so that spending the
occupancy budget on appearance does not distort overall shape. This guardrail
is a \emph{Stage-A} (matched-budget voxel-completion) notion; the full-pipeline
and building-stage evaluations report perception and buildability and make no
geometric claim. The main paper's Stage-A fidelity table already reports
Chamfer distance and F-score at $\tau{=}0.02$; Table~\ref{supp:tab_geom} completes
the guardrail with normal consistency (NC) and the tighter F-score at
$\tau{=}0.01$, under the identical matched-budget protocol (mean over $198$
objects, $R\in\{24,32,48\}$, $\rho_{\mathrm{tgt}}\in\{0.40,0.62,0.75\}$). Ours
denotes the completion network with post-hoc connectivity repair.

\begin{table}[t]
\centering\small
\setlength{\tabcolsep}{5pt}
\begin{tabular}{lcccc}
\toprule
Selector & Chamfer$\downarrow$ & F@$0.01\uparrow$ & F@$0.02\uparrow$ & NC$\uparrow$ \\
\midrule
Top-K          & \textbf{0.037} & 0.194 & \textbf{0.654} & \textbf{0.869} \\
Dilate           & 0.038 & 0.194 & 0.653 & \textbf{0.869} \\
6-sep            & 0.040 & \textbf{0.205} & 0.645 & 0.787 \\
26-sep           & 0.045 & 0.197 & 0.610 & 0.777 \\
\textbf{Ours}    & 0.038 & 0.187 & 0.641 & \textbf{0.869} \\
\midrule
\textit{TSDF (GT depth)}$^{\dagger}$ & \textit{0.030} & \textit{0.242}
& \textit{0.735} & \textit{0.889} \\
\bottomrule
\end{tabular}
\\[2pt]
{\footnotesize $^{\dagger}$TSDF consumes ground-truth depth: an oracle-input
upper bound, not a competing selector.}
\caption{Stage-A geometry guardrail under a matched occupancy budget
(mean over $198$ objects, $R\in\{24,32,48\}$,
$\rho_{\mathrm{tgt}}\in\{0.40,0.62,0.75\}$). Ours matches the strongest
geometric selector on every geometry metric while winning perception.
TSDF consumes ground-truth depth and is an oracle-input upper bound, not a competing selector; it is
marked with $\dagger$ for reference. Best among selectors in \textbf{bold}.}
\label{supp:tab_geom}
\end{table}

Across all four geometry metrics Ours sits in the top group of learned/geometric
selectors---tying the best on Chamfer ($0.038$) and NC ($0.869$) and staying
within the F-score cluster---confirming the main-paper statement that our
completion \emph{matches the strongest selector on the geometry guardrail while
winning every perceptual metric}. The only method with notably lower Chamfer / higher
F-score is TSDF, which is fed ground-truth depth and is therefore an upper bound
rather than a competitor. In short, budgeting occupancy for appearance does not
sacrifice geometric fidelity.


\subsection{Buildability Hyperparameters}
\label{supp:hparams_build}

\subsubsection{Search protocol and selection criterion}
All building-stage weights were selected on the independent
sweep-validation set (\texttt{sweepval3}).

Each configuration is judged on two tiers. The distilled stability
surrogate (Sec.~\ref{supp:surrogate_arch}, ${\sim}0.02$s/object on CPU,
${\sim}1000{\times}$ faster than the exact solver) sweeps the full grid to
reveal trends, and the exact Gurobi force-balance solver then re-certifies
the default cell plus the per-metric optima ($15$ cells). Because the
surrogate's deviation from the solver is one-sided and conservative
(Sec.~\ref{supp:surrogate_results}), grid \emph{orderings} are reliable
while absolute stability values are read only from the solver.

Because the voxel grid is coarse, weights are swept on a roughly
logarithmic scale with the default placed centrally. The building-stage
ranges are: $w_{\mathrm{area}}\in\{5,15,45,90\}$,
$w_{\mathrm{support}}\in\{5,20,60,120\}$,
$\lambda_{\mathrm{int}}\in\{0,4,12,36\}$,
$\lambda_{\mathrm{edge}}\in\{0,1,4,12\}$, and
$w_{\mathrm{col}}\in\{0,5,10,20,40,80\}$, with the lateral-neighbor
threshold fixed at $k{=}2$. The selection criterion is: attain full
certified stability on sweep-val while keeping brick count low, then take
the most color-faithful setting that does not sacrifice stability.

\subsubsection{Sensitivity: weight sweeps}
\begin{figure*}[t]
\centering
\includegraphics[width=\textwidth]{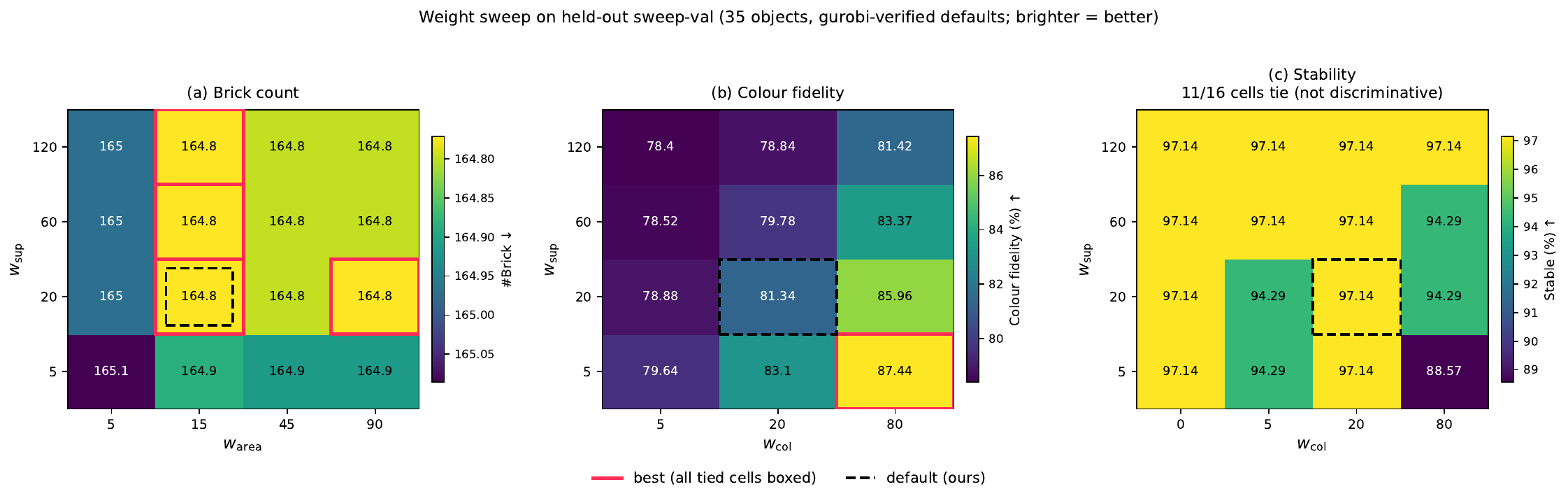}
\caption{Building-stage weight sensitivity on the $35$-object sweep-val
set (exact-solver--corrected). \textbf{Left:} brick count over
$w_{\mathrm{area}}\times w_{\mathrm{support}}$; the default sits on the
low-brick-count plateau. \textbf{Middle/right:} the
$w_{\mathrm{col}}\times w_{\mathrm{support}}$ interaction, showing color
fidelity (middle, $\uparrow$) and certified stability (right, $\uparrow$):
higher $w_{\mathrm{col}}$ improves color but needs a matching
$w_{\mathrm{support}}$ to stay fully stable. In each panel the solid box
marks the per-panel optimum and the dashed box marks the default
configuration; the default is at or adjacent to the optimum throughout.}
\label{supp:fig_sweep}
\end{figure*}

The sweeps confirm the default is at or beside the optimum on every axis,
and expose one non-trivial interaction. On
$w_{\mathrm{area}}\times w_{\mathrm{support}}$ (Fig.~\ref{supp:fig_sweep},
left) brick count is flat near the default $(15,20)$ and stability is at
its plateau maximum, so buildability weights are not brittle. On
$\lambda_{\mathrm{int}}\times\lambda_{\mathrm{edge}}$ the differences are
within noise (stability constant across the grid), indicating the
look-ahead term is not load-bearing for stability; we therefore omit its
heatmap. The key interaction is
$w_{\mathrm{col}}\times w_{\mathrm{support}}$
(Fig.~\ref{supp:fig_sweep}, middle and right): raising $w_{\mathrm{col}}$
monotonically improves color fidelity but, in isolation, erodes stability
(e.g.\ $w_{\mathrm{col}}{=}80$ with $w_{\mathrm{support}}{=}5$ drops
Stab\% to $88.6$); pairing it with a larger $w_{\mathrm{support}}$
restores full stability. Color and support must therefore be tuned
together rather than treated as independently additive; the
$w_{\mathrm{col}}\times w_{\mathrm{area}}$ sweep shows the same trend and
is omitted for brevity.

The default weights (Table~\ref{supp:tab_hparams_build}) are the unique
exact-solver--certified cell that is simultaneously fully stable
($n_u{=}0$, Stab\%${=}100$ on the $35$ sweep-validation objects) and among
the most color-faithful; since only $w_{\mathrm{col}}{=}80$ actually breaks
stability, the default $w_{\mathrm{col}}{=}20$ sits well inside the safe
region.

\begin{table}[t]
\centering\small
\setlength{\tabcolsep}{5pt}
\begin{tabular}{ll}
\toprule
Hyperparameter & Value \\
\midrule
\multicolumn{2}{l}{\emph{Building stage --- greedy scoring weights}} \\
$w_{\mathrm{area}}$       & $15$ \\
$w_{\mathrm{support}}$    & $20$ \\
$\lambda_{\mathrm{int}}$  & $12$ \\
$\lambda_{\mathrm{edge}}$ & $4$ \\
$w_{\mathrm{col}}$        & $20$ \\
lateral-neighbor threshold $k$ & $2$ \\
\bottomrule
\end{tabular}
\caption{Final building-stage greedy scoring weights, selected on the
$35$-object sweep-validation set and certified by the exact
force-balance solver.}
\label{supp:tab_hparams_build}
\end{table}

\subsection{Color Term: Component Decomposition}
\label{supp:color_ablation}
The main-paper $w_{\mathrm{col}}$ sweep ($0/20/80$;
Sec.~\ref{supp:hparams_build}) establishes that the color term helps overall.
Here we decompose the term's two factors---local isolation $\mathrm{imp}(v)$
and global rarity $\mathrm{rar}(c)$ of Eq.~\eqref{eq:colorweight}---to answer
whether each is necessary or merely decorative. All runs use the same
independent $35$-object sweep-validation set as Sec.~\ref{supp:hparams_build},
with stability certified by the exact Gurobi solver.

\paragraph{Two color metrics.}
Because a brick is single-colored, it is first assigned a representative color
by the importance-weighted vote of Eq.~\eqref{eq:cbest},
$c^\star(b)=\arg\max_c\sum_{v\in b,\,c(v)=c}\mathrm{imp}(v)$. We then measure,
over \emph{all} colored surface voxels (occupied with a valid color id;
colorless interior voxels are excluded), whether each voxel's target color
matches the representative color of its brick $b(v)$:
\begin{align}
\text{hit\%}&=100\cdot\frac{|\{v: c(v)=c^\star(b(v))\}|}{|\{v\}|},\\
\text{weighted\%}&=100\cdot\frac{\sum_{v:\,c(v)=c^\star(b(v))}\mathrm{imp}(v)}
{\sum_v \mathrm{imp}(v)}.
\end{align}
The unweighted \emph{hit\%} gives every voxel equal weight and is therefore
dominated by the large uniform color regions (which contain most voxels),
measuring whether the overall color blocks are right. The \emph{weighted\%}
weights by $\mathrm{imp}(v)$ (Eq.~\eqref{eq:imp}, $1$ minus the same-color
neighbor fraction), so isolated texture voxels count heavily while the interior
of a uniform region has $\mathrm{imp}\!\approx\!0$ and barely scores; it thus
measures whether the \emph{key texture} is preserved. Crucially, when comparing
variants $\mathrm{imp}$ is always computed under the \emph{full} definition
(the judge does not change with the variant), so the numbers are comparable
across rows rather than each variant grading itself.

\paragraph{Results.}
Table~\ref{supp:tab_color_ablation} reports the decomposition. Removing the
color term entirely (no-color) costs $2.48$ hit-\% ($p{=}1{\times}10^{-4}$) and
$3.64$ weighted-\% ($p{=}4.5{\times}10^{-6}$) against the full term, at a cost
of $+7.4$ bricks ($+4.5\%$) and \emph{zero} stability cost (both
$n_u{=}0.00$, Stab\%${=}100$). Removing only the rarity factor (no-rarity)
drops weighted fidelity by $1.71$ points ($p{=}0.024$) while barely moving
hit\% ($+1.02$, not significant, $p{=}0.25$): rare colors are exactly the
few ``minority but high-$\mathrm{imp}$'' voxels, so dropping them hardly
changes the equal-weight hit rate but clearly hurts the weighted metric---the
rarity factor protects precisely these voxels, matching its design intent. As
a cross-check, the full variant's brick count ($172.1$) is identical to the
$w_{\mathrm{col}}{=}20$ cell of Sec.~\ref{supp:hparams_build}, so the two
independent code paths agree.

\begin{table}[t]
\centering\small
\setlength{\tabcolsep}{4pt}
\begin{tabular}{lccccc}
\toprule
Variant & weight $w(v)$ & \#Br & hit\% & wtd\% & Stab\% \\
\midrule
\textbf{full (ours)} & $\mathrm{imp}\times\mathrm{rar}$ & $172.1$ & $82.90$ & $81.34$ & $\mathbf{100.0}$ \\
no-rarity & $\mathrm{imp}\times 1$ & $162.8$ & $81.88$ & $79.62$ & $100.0$ \\
no-color & $w_{\mathrm{col}}{=}0$ & $164.8$ & $80.42$ & $77.70$ & $100.0$ \\
\midrule
no-importance$^{\dagger}$ & $1\times\mathrm{rar}$ & $175.2$ & $84.03$ & $82.50$ & $97.1$ \\
\bottomrule
\end{tabular}
\\[2pt]
{\footnotesize $^{\dagger}$Reported for completeness only, \emph{not} read as
``importance is harmful.'' Both color metrics assign the brick's
representative color by an \emph{importance}-weighted vote, so a variant that
drops $\mathrm{imp}$ at brick-selection time yet is still graded under an
$\mathrm{imp}$-weighted judge is favored by a self-referential bias. Its higher
brick count ($175.2$, $+3$ over full) trades more bricks for fidelity rather
than being a better operating point, and it is the only variant that breaks
stability ($n_u{=}0.03$, Stab\%${=}97.1$). Visual quality is adjudicated by the
colored-render perceptual metrics, which use the target colored voxels as
reference and have no such self-reference.}
\caption{Color-term component decomposition on the $35$-object
sweep-validation set (all stability exact-Gurobi certified). \#Br: brick
count; hit\%/wtd\%: unweighted / importance-weighted color-fidelity
(defined above); Stab\%: fully stable models. Paired Wilcoxon ($n{=}35$)
full vs.\ no-color: hit $+2.48$ ($p{=}1{\times}10^{-4}$), weighted $+3.64$
($p{=}4.5{\times}10^{-6}$); full vs.\ no-rarity: weighted $+1.71$
($p{=}0.024$), hit $+1.02$ (n.s., $p{=}0.25$).}
\label{supp:tab_color_ablation}
\end{table}

\paragraph{Conclusions.}
(i)~The color term is necessary: it improves color hit rate by $2.48$ points
and weighted fidelity by $3.64$ points over no-color at a modest $+4.5\%$ brick
cost and \emph{zero} stability cost. (ii)~The rarity factor contributes: without
it weighted fidelity drops $1.71$ points ($p{=}0.024$), i.e.\ rare colors are no
longer preferentially protected; the hit-rate change is small and insignificant
precisely because rarity acts on the few key voxels the weighted metric is
designed to capture. The $\mathrm{imp}$-only (no-importance) row is reported only
for completeness and is discounted for the self-referential reason noted in the
table.

\section{Analysis and Scope of Claims}
\subsection{On the Scope of Buildability Claims}
\label{supp:scope}
Our method touches four distinct notions of ``buildability'' that are easy to
conflate. We separate them here and state, for each, whether it is a
method-level \emph{guarantee} or an \emph{empirical} result, so that no
stronger claim is read into the paper than we make.

\paragraph{(1) Grounded connectivity --- guaranteed by construction.}
This is our only method-level guarantee. The building stage repairs every
assembly to a single ground-anchored component: a union--find pass flags each
floating component, and the two-tier repair (zero-deformation recombination,
then cap/shelf bridging, with removal as a last resort) reduces the floating
count at every atomic step, so the assembly reaches full ground connectivity in
finitely many steps (main paper, Building Stage). The output therefore contains
\emph{zero floating components} by construction, independent of the input
object. This is a structural (graph-connectivity) property, not a physical
force-balance guarantee.

\paragraph{(2) Local support --- soft objective, not guaranteed.}
The support term in the greedy placement score rewards bricks resting on solid
support below and adds a rescue bonus for bricks that anchor otherwise
unsupported cells. This biases placement toward well-supported layouts but is a
\emph{soft preference} optimized during scoring, not a constraint: individual
unsupported cells may remain, and the fraction of unsupported cells is reported
empirically, never claimed to be zero.

\paragraph{(3) Exact static stability --- empirically certified, not a theorem.}
Static stability is a physical property (does the assembly stand under gravity
and contact forces) that our construction does \emph{not} prove. Every reported
stability number (\emph{Stab\%}) is \emph{certified post hoc} by the exact
Gurobi force-balance solver on the finished assembly; a distilled
message-passing surrogate (per-brick AUC $0.984$) is used only as a fast in-loop
screen and never contributes to reported numbers. Consequently the $100\%$ in
the main results table is an \textbf{empirical fraction measured on the test
set}, not a method-level theorem: it states that every certified assembly in
that evaluation passed the exact solver, not that the method can never produce
an unstable assembly.

\paragraph{(4) Physical hand-buildability --- empirical.}
Whether the exported instructions can actually be assembled by a person is
validated empirically by the four real-world builds. The four objects
(\texttt{fire\_extinguisher\_010}, \texttt{toy\_plant\_001},
\texttt{toy\_boat\_010}, and \texttt{timer\_005}, all from
OmniObject3D~\cite{omniobject3d}) span diverse categories and are built by hand
directly from the pipeline's exported layer-by-layer instructions without
manual edits, each standing unsupported on a flat surface (main paper,
real-world validation). This is an existence-style empirical demonstration on a
handful of objects, not a guarantee for arbitrary inputs.

\paragraph{Summary and scope.}
Only grounded connectivity (1) is guaranteed by construction; local support (2)
is a soft objective, and exact static stability (3) and physical
hand-buildability (4) are empirical, certified respectively by the exact solver
and by real builds. This matches our two-stage framing: the completion stage
allocates the occupancy budget for appearance under a connectivity constraint,
and the building stage delivers grounded connectivity by construction while
stability is certified rather than proven. 
We deliberately do not fold a brick-level stability or force-balance objective into the occupancy selector: doing so would force every intermediate build state to be self-standing, and thereby ignore the cap support that later overlapping layers can provide. Such a constraint can render otherwise-buildable shapes locally infeasible, such as bridge- or arch-like structures, which are transiently unsupported mid-assembly yet fully grounded once the spanning layers close over them. Coupling the two stages into a single joint optimization that reasons about this cross-layer support is left as future work, so we make no joint-optimization guarantee here.

\subsection{Lazy-Greedy Oracle Construction: Justification}
\label{supp:celf}
The perceptual completion oracle (main paper, Stage-A oracle objective:
depth $+$ silhouette $+$ normal $+$ LPIPS terms rendered through the
reconstruction) is built by a \emph{CELF-style lazy-greedy}
selector~\cite{leskovec2007celf}: at each step only the candidates whose
local neighborhood changed are re-scored, using a gain-ordered frontier
with batched six-view rendering. We clarify here what is and is not
claimed by this choice.

\paragraph{We do not claim submodularity.}
Strict CELF is provably identical to exhaustive forward greedy \emph{when
the objective is monotone submodular}, because a stale (previously
computed) marginal gain is then an upper bound on the current gain, so a
freshly re-scored top-of-frontier candidate is provably the argmax. Our
objective mixes a deep perceptual distance (LPIPS) with a compositional
rendering and is \emph{not} known to be submodular; we therefore make no
submodularity claim and invoke no $1-1/e$ optimality guarantee. CELF-style
lazy evaluation is used purely as an \emph{accelerator} that reproduces the
plain-greedy selection while avoiding redundant re-scoring, not as an
exactness theorem about our loss.

\paragraph{The shipped selector reproduces strict CELF exactly.}
Our production selector re-scores the top-$M$ ($M{=}48$) stalest frontier
candidates in a single batched pass before committing the argmax, rather
than popping one at a time. To confirm this batching introduces no
deviation, we compared it against strict one-at-a-time CELF and against
exhaustive greedy on sweepval over 3 resolutions($R=(24/32/48)$), with candidate pools $N_b$ from $128$ to $1254$. The
batched selector is \textbf{index-for-index identical to strict CELF on
every object} (selection overlap $=1.000$, exactly), so no approximation
is introduced by batching.

\paragraph{Greedy tracks exhaustive within a negligible loss gap.}
Against exhaustive greedy, the lazy selector's chosen \emph{indices}
diverge (mean overlap $0.77$)---direct evidence that the objective is not
strictly submodular, since a submodular objective would force overlap
$1.0$. Yet the resulting \emph{loss} is essentially unchanged: the final
objective gap versus exhaustive is $+0.6\%$ on average (worst $+2.0\%$),
and on two objects the lazy selection is marginally \emph{better} than
exhaustive ($-0.0003$), a signature of mild non-monotonicity / near-ties.
The objective thus behaves as \emph{approximately} submodular in our
operating regime---greedy picks a different but equally good set---though
we do not measure a submodularity ratio and make no formal approximation
claim beyond this empirical bracket.

\paragraph{The oracle is a teacher, not an inference-time selector.}
Crucially, this greedy construction is run \emph{offline} to generate
training targets only: the completion network is distilled against the
oracle's per-voxel ranking (Phase~1), and at inference we take the
top-$n_{\mathrm{occ}}$ scored voxels from the network in a single forward
pass---no greedy search is run at test time. The oracle's role is to
provide a strong supervision signal, and it is empirically the strongest
selector we have: on the held-out set the oracle-distilled network
outperforms uniform voxelization and the budgeted top-$K$/dilation
selectors on the perceptual metrics (main paper, Stage-A fidelity table).
Any residual sub-optimality in the greedy target is therefore absorbed
by distillation rather than propagated to inference.

\subsection{Concurrent Systems Not Run as Quantitative Baselines}
\label{supp:concurrent}
Two recent systems are closely related in spirit but are not included as
quantitative baselines, for reasons of both availability and task setup.

\textit{BrickAnything}~\cite{brickanything} is the closest in objective
(explicit 3D-geometry input, appearance fidelity, and buildability).
We do not compare against it numerically for three reasons:
(i) its setup differs from ours---it consumes an explicit
point-cloud/3D-geometry input and drives buildability through a
reward-driven autoregressive prior, rather than reconstructing appearance
from pose-free images under an occupancy budget;
(ii) neither public code nor trained weights are available, so it cannot
be re-run on our unified data under a matched budget; and
(iii) it is a contemporaneous preprint. We therefore position it
qualitatively in the related work rather than reporting numbers, to avoid
an unequal comparison on a setup we cannot faithfully reproduce.

\textit{DVD}~\cite{dvd} (discrete voxel diffusion) targets a different
problem: it generates or edits voxel volumes on a \emph{fixed} $64^3$
lattice conditioned on image or text, with no notion of a variable
occupancy budget and no brick-level buildability objective. Its task
formulation is thus not directly comparable to our budgeted, buildability
aware, pose-free reconstruction setting; we cite it as related generative
prior work rather than as a baseline.

\section{Limitations and Future Directions}
\label{supp:limitations}
As stated in the main paper's conclusion, several limitations bound the
current method and point to natural extensions; we expand on them here.

\paragraph{Intrinsic ceiling of low-resolution occupancy.}
Our setting deliberately targets a coarse lattice under a tight occupancy
budget, which is what makes visual resemblance and buildability pull against
each other in the first place. This coarse-lattice regime is a design choice
rather than a defect---it matches how physical brick models are actually
built---but it inherently caps the expressible detail: features finer than a
single cell cannot be recovered at a given resolution regardless of how the
budget is allocated. As noted in the main paper, raising the resolution or
adaptively varying the budget across an object are the direct ways to lift this
ceiling, at the cost of the exact solver's and the buildability baselines'
tractability, which is precisely why our comparisons are conducted at the
coarsest resolution.

\paragraph{Two-stage design rather than joint optimization.}
We treat occupancy completion and brick assembly as two stages: the completion
network allocates the budget for appearance under a connectivity constraint,
and the building stage delivers grounded connectivity by construction with
stability certified post hoc. As discussed in the scope of our claims
(Sec.~\ref{supp:scope}), we do not fold a brick-level stability or brick-count
objective into the occupancy selector, so the two stages are not jointly
optimized. This keeps each stage simple and analyzable and is sufficient
because the budget allocated in the first stage upper-bounds what any assembly
can achieve; tightening the coupling---e.g.\ propagating a differentiable
brick-count or force-balance signal back into completion---is a promising
direction for further improving the parsimony--stability balance.

\paragraph{Dependence on the front-end reconstruction.}
Our contribution is downstream of the recovered mesh: we take a watertight
textured mesh from a pose-free reconstructor (FreeSplatter in our
implementation) and convert it into a buildable brick model. As stated in the
main paper, any comparable reconstructor that yields such a mesh could
substitute for the front end, but errors in that upstream stage (missing thin
structures, texture bleeding, or geometry artifacts) propagate into the brick
model, since our method does not attempt to correct the input geometry. Coupling
the brick-conversion objective with the reconstruction front end, so that
downstream buildability and fidelity can inform the recovered geometry, is left
for future work.

\end{document}